\documentclass[sigconf]{acmart}

\AtBeginDocument{%
  }

\acmISBN{978-1-4503-XXXX-X/2018/06}
\usepackage{times}
\usepackage{latexsym}
\usepackage{algorithm}
\usepackage{algpseudocode}
\usepackage{booktabs}
\usepackage{multicol}
\usepackage{multirow}
\usepackage{float}
\usepackage{tcolorbox}
\usepackage[T1]{fontenc}
\usepackage[utf8]{inputenc}
\usepackage{microtype}
\usepackage{graphicx}

\usepackage{amssymb}
\usepackage{xcolor}
\usepackage{tablefootnote}
\definecolor{darkgreen}{RGB}{0, 100, 0}

\begin{document}

\title{Hierarchical Prompting Taxonomy: A Universal Evaluation Framework for Large Language Models Aligned with Human Cognitive Principles}

\author{Devichand Budagam}
\affiliation{%
  \institution{Indian Institute of Technology Kharagpur}
  \country{India}
}
\author{Ashutosh Kumar}
\affiliation{%
  \institution{Rochester Institute of Technology}
  \country{USA}
}

\author{Mahsa Khoshnoodi}
\affiliation{%
  \institution{Researcher, Fatima Fellowship}
  \country{USA}
}

\author{Sankalp KJ}
\affiliation{%
  \institution{AI Institute, University of South Carolina}
  \country{USA}
}

\author{Vinija Jain}
\authornote{Work does not relate to position at Amazon.}
\affiliation{%
  \institution{Amazon GenAI}
  \institution{Stanford University}
  \country{USA}
}

\author{Aman Chadha}
\authornotemark[1]
\affiliation{%
  \institution{Amazon GenAI}
  \institution{James Silberrad Brown Center for AI, San Diego State University}
  \institution{Stanford University}
  \country{USA}
}


\begin{abstract}
 Assessing the effectiveness of large language models (LLMs) in performing different tasks is crucial for understanding their strengths and weaknesses. This paper presents Hierarchical Prompting Taxonomy (HPT), grounded on human cognitive principles and designed to assess LLMs by examining the cognitive demands of various tasks. The HPT utilizes the Hierarchical Prompting Framework (HPF), which structures five unique prompting strategies in a hierarchical order based on their cognitive requirement on LLMs when compared to human mental capabilities. It assesses the complexity of tasks with the Hierarchical Prompting Index (HPI), which demonstrates the cognitive competencies of LLMs across diverse datasets and offers insights into the cognitive demands that datasets place on different LLMs. This approach enables a comprehensive evaluation of LLM’s problem-solving abilities and the intricacy of a dataset, offering a standardized metric for task complexity. Extensive experiments with multiple datasets and LLMs show that HPF enhances LLM performance by \(2 \to 63 \%\) compared to baseline performance, with GSM8k being the most cognitively complex task among reasoning and coding tasks with an average HPI of \(3.20\) confirming the effectiveness of HPT. To support future research in this domain, the implementations of HPT and HPF are publicly available\footnote{\href{https://github.com/devichand579/HPT}{Code and Experiments}}.
\end{abstract}







\maketitle
\pagestyle{plain}

\section{Introduction}
\label{sec: Introduction}
Large Language Models (LLMs) have revolutionized natural language processing (NLP), enabling significant advancements in a wide range of applications. Conventional evaluation frameworks often apply a standard prompting approach to assess different LLMs, regardless of the complexity of the task, which may result in biased and suboptimal outcomes.  Moreover, applying the same prompting approach across all samples within a dataset without considering each sample's relative complexity adds to the unfair situation. To achieve a more balanced evaluation framework, it is essential to account for both the task-solving ability of LLMs and the varying cognitive complexities of the dataset samples. This limitation highlights the need for more sophisticated evaluation methods that can adapt to varying levels of sample task complexity.
\begin{figure}
  \centering
  \includegraphics[width=0.7\linewidth]{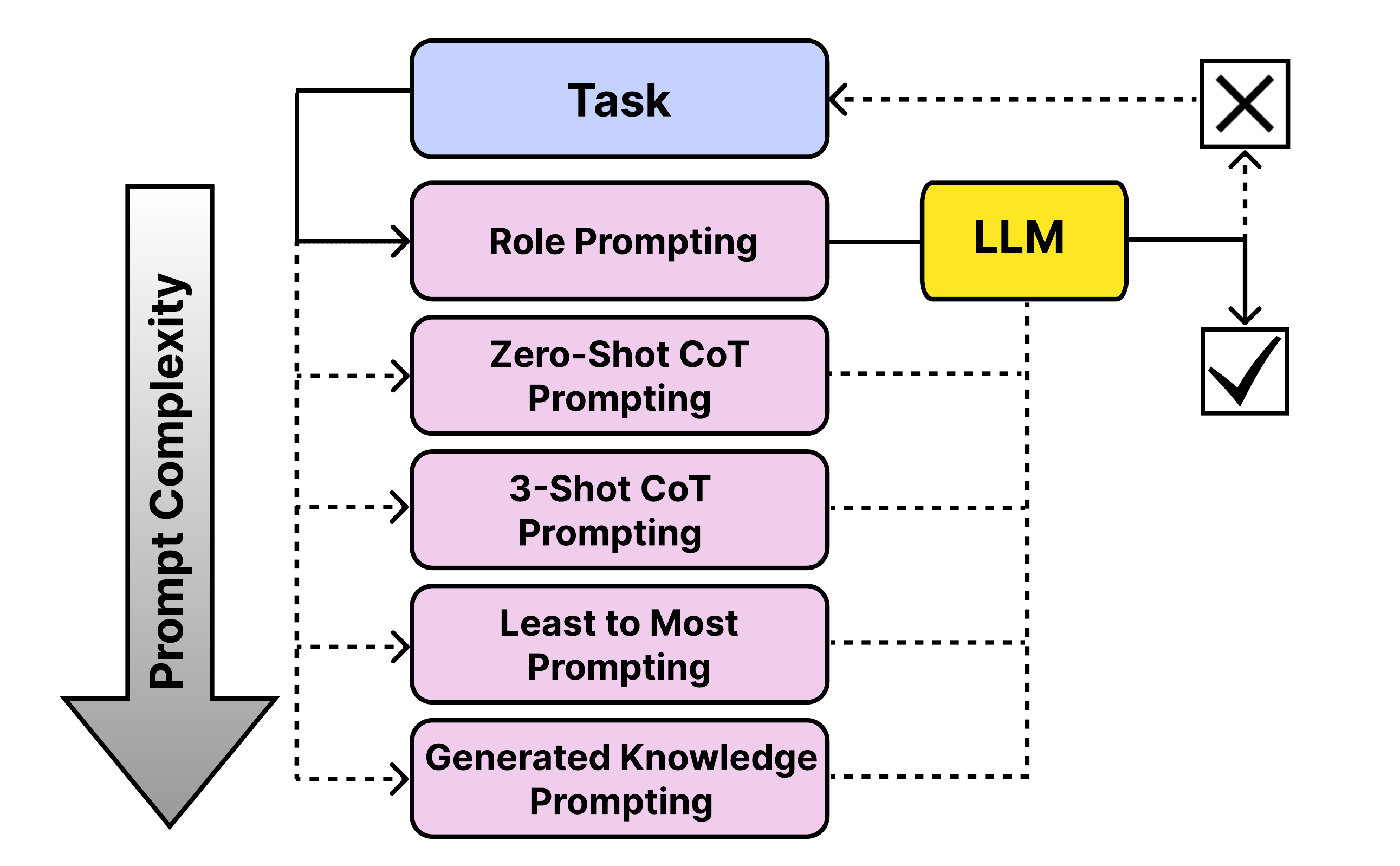}
  \Description{Hierarchical Prompting Framework diagram.}
  \caption{The Hierarchical Prompting Framework includes five distinct prompting strategies, each designed for different levels of task complexity to ensure the appropriate prompt is selected for the given task. A $\checkmark$ indicates task completion, while a $\times$ signifies task incompletion.}
  \label{fig:1}
\end{figure}
This study defines \emph{complexity} as the cognitive demands imposed by a task or the cognitive load introduced by a prompting strategy on LLMs. Task complexity in human cognition reflects the mental effort required for processing, analyzing, and synthesizing information. As \citet{SWELLER1988257} noted, complexity increases with greater cognitive resource demands, engaging working memory in reasoning and problem-solving. Similarly, \citet{anderson2014taxonomy} describes cognitive abilities as a continuum, from basic recall to higher-order thinking, with difficulty rising for tasks requiring analysis, synthesis, and evaluation. By mapping LLM prompting strategies onto this hierarchy, we systematically assess how LLMs handle varying cognitive loads. This framework provides a structured, cognitively grounded method for evaluating model performance across tasks of differing complexity. This study is directed by the following research questions:
\begin{tcolorbox}[
    colback=purple!15,    
    colframe=purple!85!black,  
    sharp corners,
    boxrule=1pt,
    width=\linewidth,
    title={\textbf{Research Questions:}},
    fonttitle=\bfseries\large
]

\begin{itemize}
   \item \textbf{RQ1:} Does cognitively inspired strategic selection of prompts enable small language models (SLMs) to match the performance of LLMs?
   \item \textbf{RQ2:} How can cognitive demand measurements of LLMs guide model selection and deployment decisions beyond traditional metrics?
   \item \textbf{RQ3:} How can we align prompt complexity with task demands to optimize both computational efficiency and performance? 
\end{itemize}
\end{tcolorbox}
This paper introduces the HPT, a set of rules that maps the human cognitive principles for assessing the complexity of different prompting strategies. It employs the HPF shown in Figure \ref{fig:1}, a prompt selection framework that selects the prompt with the optimal cognitive load on LLM required in solving the task.
The main contributions of this work are:
\begin{itemize}
   \item \textbf{Hierarchical Prompting Taxonomy (HPT):} The paper introduces HPT, rules mapping prompting strategies to human cognitive principles, enabling a universal measure of LLMs' task complexity.

\item \textbf{Hierarchical Prompting Framework (HPF):} The HPF framework selects the best prompt from five strategies to optimize LLMs' cognitive load, improving evaluation and performance transparency.

\item \textbf{Hierarchical Prompting Index (HPI):} HPI\footnote{HPI can be quantitatively assessed to analyze the cognitive abilities of an LLM and the cognitive demands imposed by datasets on LLMs, as both factors are interchangeably related to the complexity of tasks.} quantitatively assesses LLMs' task complexity across datasets, revealing cognitive demands on various LLMs.
\end{itemize}
\begin{figure}[h]
    \centering
    \includegraphics[width=0.9\linewidth]{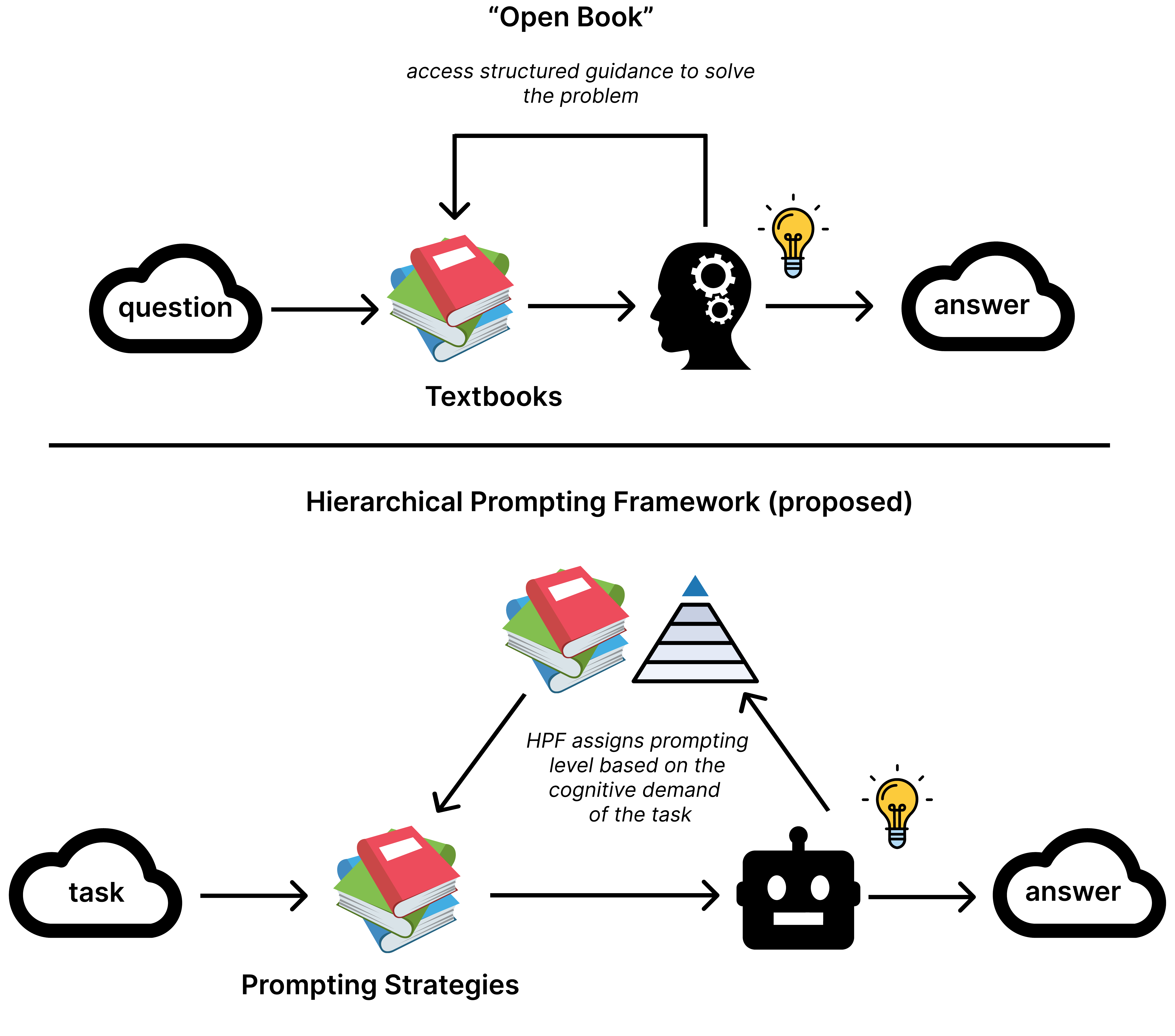}
    \caption{Analogical framework comparing the HPF with "open book" examination methodology. The diagram illustrates how HPF components (below) mirror traditional educational assessment elements (above), with parallel relationships between task complexity levels, resource utilization (prompts/textbooks), and performance metrics (HPI/student effort). This comparison demonstrates how LLM task complexity scales similarly to educational assessment complexity, from simple lookup tasks to complex synthesis problems.}
    \label{fig:analogy}
\end{figure}
HPF can be compared to an \textit{"open book"} exam (see Figure \ref{fig:analogy}), with tasks analogous to questions and prompting strategies akin to textbooks. The exam questions, ranging from basic recall to complex analysis, parallel the cognitive challenges in HPT tasks. Similarly, textbooks offer structured support, much like HPF, which arranges prompts by complexity to assist LLMs. A glossary lookup represents a task with low complexity, whereas solving a multi-step analytical problem indicates high complexity. The effort exerted by a student is similar to HPI, which measures the cognitive demand on LLMs. Just as structured learning materials improve students' performance, carefully crafted hierarchical prompts help LLMs in addressing increasingly complex tasks more effectively.

The remainder of the paper is structured as follows: Section \ref{sec:Related Work} reviews the related work on prompting and evaluation in LLMs. Section \ref{sec:Methodology} details the HPT and its associated frameworks. Section \ref{sec:Results} outlines the experimental setup, results, and ablation studies. Section \ref{sec:Conclusion} concludes the paper. Section \ref{sec:Limitations} discusses the limitations of the work. Section \ref{sec:ethical impact} discusses the ethical impact of the work.
\section{Related Work}
\label{sec:Related Work}
The advent of LLMs has revolutionized NLP by demonstrating significant improvements in few-shot and zero-shot learning capabilities. \citet{brown2020language} introduced GPT-3, a 175 billion parameter autoregressive model, showcasing its ability to perform a wide range of tasks such as question-answering, reading comprehension, translation, and natural language inference without fine-tuning. This study highlighted the potential of very large models for in-context learning while also identifying limitations in commonsense reasoning and specific comprehension tasks. Similarly, \citet{liu2021pretrain} surveyed prompt-based learning, emphasizing the role of prompt engineering in leveraging pre-trained models for few-shot and zero-shot adaptation to new tasks with minimal labeled data.
\subsection{Prompt Engineering}
Prompting plays a vital role in unlocking the full potential of LLMs. By designing specific input prompts, the LLM's responses can be guided, significantly influencing the quality and relevance of the output. Effective prompting strategies have enhanced LLM performance on tasks ranging from simple question-answering to complex reasoning and problem-solving. Recent research has explored various approaches to prompting and reasoning evaluation in LLMs. Chain-of-Thought (CoT) prompting \cite{wei2022chain} elicits step-by-step reasoning, improving performance on complex tasks. Specializing smaller models \cite{pmlr-v202-fu23d} and using large models as reasoning teachers \cite{ho2022large} have demonstrated the potential for enhancing reasoning capabilities. Emergent abilities in LLMs, which appear suddenly at certain scale thresholds, have also been a topic of interest. \citet{wei2022emergent} examined these abilities in few-shot prompting, discussing the underlying factors and implications for future scaling. Complementing this, \citet{kojima2022large} demonstrated that LLMs could exhibit multi-step reasoning capabilities in a zero-shot setting by simply modifying the prompt structure, thus highlighting their potential as general reasoning engines. \citet{yao2023tree} introduced the Tree-of-Thoughts framework, enabling LLMs to deliberate over coherent text units and perform heuristic searches for complex reasoning tasks. This approach generalizes over chain-of-thought prompting and has shown significant performance improvements in tasks requiring planning and search, such as creative writing and problem-solving games. \citet{kong2024better} introduced role-play prompting to improve zero-shot reasoning by constructing role-immersion interactions, which implicitly trigger chain-of-thought processes and enhance performance across diverse reasoning benchmarks. Progressive-hint prompting \cite{zheng2023progressivehint} has been proposed to conceptualize answer generation and guide LLMs toward correct responses. Metacognitive prompting \cite{wang2024metacognitive} incorporates self-aware evaluations to enhance understanding abilities.

These studies highlight progress in using innovative prompting techniques to improve LLMs' emergent abilities, reasoning, interaction strategies, robustness, and evaluation. Yet, challenges persist in prompt design, managing complex reasoning tasks, and performance evaluation across various scenarios. Although LLMs show promising emergent abilities, they frequently lack predictability and control, and their resistance to misleading prompts is still an issue.
\subsection{Prompt Optimization and Selection}
The challenge of optimizing prompts for LLMs has been addressed in several key studies, each contributing unique methodologies to enhance model performance and efficiency. \citet{shen2023flatness} introduce PFLAT, a metric utilizing flatness regularization to quantify prompt utility, which leads to improved results in classification tasks. \citet{do2024automaticpromptselectionlarge} propose a structured three-step methodology that contains data clustering, prompt generation, and evaluation, effectively balancing generality and specificity in prompt selection. ProTeGi \cite{pryzant2023automatic} offers a non-parametric approach inspired by gradient descent, leveraging natural language "gradients" to iteratively refine prompts. \citet{wang-etal-2024-promise} present PromISe, which transforms prompt optimization into an explicit chain of thought, employing self-introspection and refinement techniques. \citet{zhou2023efficient} proposed DYNAICL, a framework for efficient prompting that dynamically allocates in-context examples based on a meta-controller's predictions, achieving better performance-efficiency trade-offs compared to uniform example allocation.

These studies seek to automate prompt design, reducing reliance on manual trial-and-error while improving efficiency and scalability across tasks and models. They report performance gains of 5\% to 31\% across benchmarks, highlighting the growing significance of prompt optimization. Future research directions include exploring theoretical foundations, combining optimization techniques, and differentiating task-specific from general-purpose strategies.
\subsection{Evaluation Benchmarks}
To facilitate the evaluation and understanding of LLM capabilities, \citet{zhu2024promptbench} introduced PromptBench, a unified library encompassing a variety of LLMs, datasets, evaluation protocols, and adversarial prompt attacks. This modular and extensible tool aims to support collaborative research and advance the comprehension of LLM strengths and weaknesses. Further exploring reasoning capabilities, \citet{qiao-etal-2023-reasoning} categorized various prompting methods and evaluated their effectiveness across different model scales and reasoning tasks, identifying key open questions for achieving robust and generalizable reasoning. \cite{wang2021adversarial} introduced a multitask benchmark for LLM robustness evaluation, which extends the original GLUE \cite{wang-etal-2018-glue} benchmark to assess model robustness against adversarial inputs. It incorporates perturbed versions of existing GLUE tasks, such as paraphrasing, negation, and noise, to test models' abilities with challenging data. The study highlights that despite their success on clean datasets, state-of-the-art models often struggle with adversarial examples, underscoring the importance of robustness evaluations in model development.
\section{Hierarchical Prompting Taxonomy}
\label{sec:Methodology}
\begin{figure*}[!ht]
    \centering
    \includegraphics[width=0.9\linewidth]{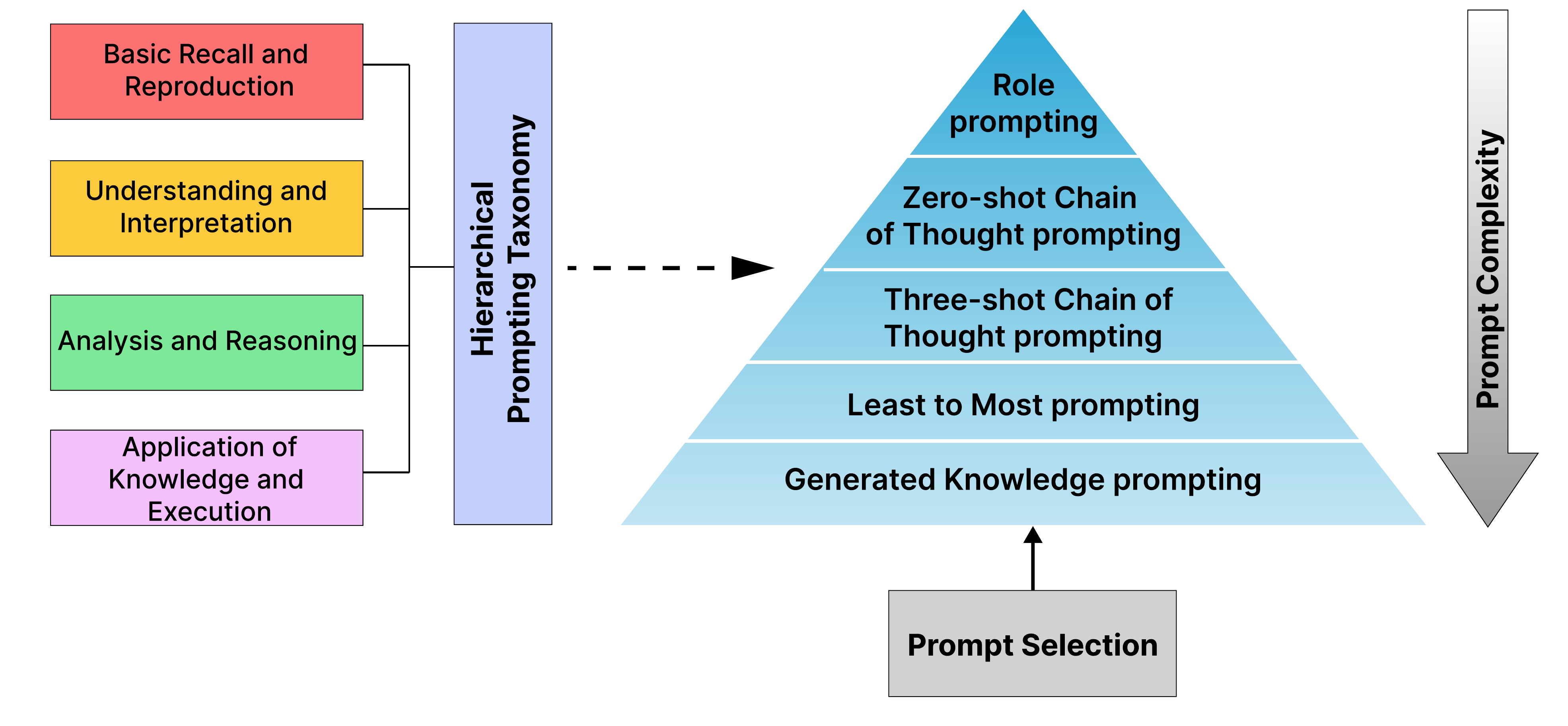}
    \caption{Hierarchical Prompting Taxonomy: A taxonomy designed to assess the complexity of prompting strategies based on the criteria: Basic Recall and Reproduction, Understanding and Interpretation, Analysis and Reasoning, and Application of Knowledge and Reasoning.}
    \label{fig:2}
\end{figure*}
\subsection{Governing Rules}
Figure \ref{fig:2} illustrates the HPT, a taxonomy that systematically reflects human cognitive functions as outlined in \citet{bloom1956taxonomy}. Each rule embodies complex cognitive processes based on established principles from learning and psychology.
\begin{enumerate}
    \item \label{1}\textbf{Basic Recall and Reproduction}: This reflects the fundamental cognitive process of remembering and reproducing factual information without analysis or interpretation, which involves mere recognition or retrieval of knowledge from memory \cite{anderson2014taxonomy}.
    \item \label{2}\textbf{Understanding and Interpretation}: This corresponds to the second cognitive rule of \cite{bloom1956taxonomy}, where individuals must not only recall information but also explain it in their own words, summarize key points or clarify the meaning of content. This rule demands an intermediate cognitive load involving information processing rather than retrieving it.
    \item \label{3}\textbf{Analysis and Reasoning}: This aligns with the analysis stage of \cite{bloom1956taxonomy}, which involves higher cognitive functions such as comparison, contrast, and deep understanding of the underlying principles. It is more complex than mere understanding because it requires examining structure and identifying patterns and connections.
    \item \label{4}\textbf{Application of Knowledge and Execution}: This mirrors the application and evaluation stages of \cite{bloom1956taxonomy}, where individuals must not only understand and analyze but also use knowledge to perform multi-step tasks, solve complex problems, and execute decisions. It represents the most cognitively complex tasks, which require synthesis of information and practical decision-making, highlighting the critical leap from understanding theory to executing it in practice.
\end{enumerate}
In HPT, the progression from basic recall to application of knowledge reflects increasing cognitive complexity, consistent with educational and cognitive frameworks, where more advanced cognitive processes build on foundational ones, demanding deeper engagement and mental effort.
\subsection{Hierarchical Prompting Framework}
The HPF consists of five prompting strategies, each assigned a complexity level. These levels are determined by the degree to which the strategies are shaped by the four principles of the HPT. The complexity levels of the prompting strategies are assigned based on human assessment of their relative cognitive loads over a set of 7 different tasks, guaranteeing that the cognitive abilities of LLMs are in harmony with those of humans. This approach enables the assessment of tasks in terms of their complexity and the cognitive load they impose on both humans and LLMs by utilizing HPI. Section \ref{sec:llm-as-a-judge} examines the hierarchical structure of the HPF in conjunction with the LLM-as-a-Judge framework, validating that the cognitive demands on LLMs can be aligned with those of humans.

The five prompting strategies were selected to ensure comprehensive coverage of cognitive demands rather than maximizing the number of strategies (see Appendix \ref{appendix: human policy}). This makes HPF adaptable, allowing for replication or expansion with similar strategies. The strategies, ordered by increasing complexity, are:
\begin{enumerate}
    \item \textbf{Role Prompting} \cite{kong2024better}: Specifies the LLM’s role in task resolution, exerting minimal influence from HPT principles.
    \item \textbf{Zero-Shot Chain-of-Thought Prompting (Zero-CoT)} \cite{kojima2022large}: Uses “Let’s think step by step” to encourage reasoning, moderately influenced by rule \ref{3}.
    \item \textbf{Three-Shot Chain-of-Thought Prompting (3-CoT)} \cite{wei2022chain}: Provides three examples to guide reasoning, strongly influenced by rules \ref{1} and \ref{2}, with moderate influence from rule \ref{3}.
    \item \textbf{Least-to-Most Prompting}  \cite{zhou2023leasttomost}: Breaks tasks into sub-problems, requiring recall, interpretation, and analysis, exerting strong influence from rules \ref{1}, \ref{2}, \& \ref{3}.
    \item \textbf{Generated Knowledge Prompting (GKP)} \cite{liu-etal-2022-generated}: Integrates external knowledge, demanding correlation, application, and analysis, making it the most cognitively complex (rules \ref{2}, \ref{3}, and \ref{4}). Llama-3 8B generates the external knowledge in experiments.
\end{enumerate}
\subsection{Hierarchical Prompting Index}
    HPI is an evaluation metric for assessing the task complexity of LLMs over different datasets, which is influenced by the HPT rules. A lower HPI for a dataset suggests that the corresponding LLM is more adept at solving the task with fewer cognitive processes. For each dataset instance, we begin with the least complex prompting strategy and progressively move through the HPF prompting strategies until the instance is resolved. The HPI corresponds to the complexity level of the prompting strategy where the LLM first tackles the instance.
\begin{algorithm}
\caption{HPI Computation}
\label{alg: HPI}
\begin{algorithmic}
    \State $\texttt{HPI\_List} = [\hspace{2pt}]$  
    \For{sample $i$ in the evaluation dataset} 
        \For{level $x$ in the HPF}
           \If{LLM resolves the task}
              \State $\texttt{HPI\_List}[i] = x$  
              \State \textbf{break}  
           \EndIf
         \EndFor
         \If{LLM failed to resolve the task}
            \State $\texttt{HPI\_List}[i] = m + \texttt{HPI}_{Dataset}$
         \EndIf
    \EndFor
    \State  $\texttt{HPI} = \frac{1}{n} \sum_{j=1}^{n} \texttt{HPI\_List}[j]$
\end{algorithmic}
\end{algorithm}
\\
Algorithm \ref{alg: HPI} illustrates the process for determining HPI, with $m$ indicating the total levels within the HPF and $n$ representing the number of samples in the evaluation dataset. $\texttt{HPI}_{Dataset}$ denotes the penalty that human evaluations impose on the framework. Additional information regarding human annotation is provided in Appendix \ref{appendix: human policy}.

\section{Results}
\label{sec:Results}
\subsection{Experimental Setup}
\textbf{Datasets}\\
We evaluated the framework on diverse datasets spanning reasoning, coding, mathematics, question-answering, summarization, and machine translation. For dataset sizes, see Appendix \ref{appendix: human policy}.\\  
\textbf{Reasoning}: MMLU \cite{hendryckstest2021} (57 subjects, multiple-choice), CSQA \cite{talmor2019commonsenseqa} (~12K commonsense questions).\\  
\textbf{Coding}: HumanEval \cite{chen2021evaluating} (164 function-based coding tasks).\\  
\textbf{Mathematics}: GSM8k \cite{cobbe2021gsm8k} (8.5K multi-step math problems).\\  
\textbf{Question-Answering}: BoolQ \cite{clark-etal-2019-boolq} (~16K True/False questions from Wikipedia).\\  
\textbf{Summarization}: SamSum \cite{gliwa-etal-2019-samsum} (~16K human-annotated dialogue summaries).\\  
\textbf{Machine Translation}: IWSLT-2017 en-fr \cite{cettolo-etal-2017-overview} (TED Talk parallel corpus).\\[5 pt]  
\textbf{Large Language Models}: 
We tested LLMs ranging from 7B to 12B parameters across open-source and proprietary models.\\  
\textbf{Proprietary LLMs}: GPT-4o \cite{openai2024gpt4o}, Claude 3.5 Sonnet \cite{anthropic2024claude35sonnet}.\\  
\textbf{SLMs}: Gemma 7B \cite{gemmateam2024gemma}, Mistral 7B \cite{jiang2023mistral}, Llama-3 8B \cite{llama3modelcard}, Gemma-2 9B \cite{team2024gemma}, Mistral-Nemo 12B \cite{mistral2024nemo}.\\[5pt]
\textbf{Additional Evaluation Metrics}\\  
\textbf{Coding}: Pass@k \cite{chen2021evaluatinglargelanguagemodels} estimates the probability of at least one correct solution among the top k outputs for code generation.\\  
\textbf{Summarization}: ROUGE-L \cite{lin-2004-rouge} measures sequence-level similarity via the longest common subsequence.\\  
\textbf{Machine Translation}: BLEU \cite{10.3115/1073083.1073135} evaluates n-gram precision against reference texts.\\[5pt]  
Summarization and translation tasks used thresholds of 0.15 and 0.20, respectively, to define task completion at each HPF complexity level, enabling iterative refinement of prompting strategies.  
\begin{figure}[]
    \centering
    \includegraphics[width=\linewidth]{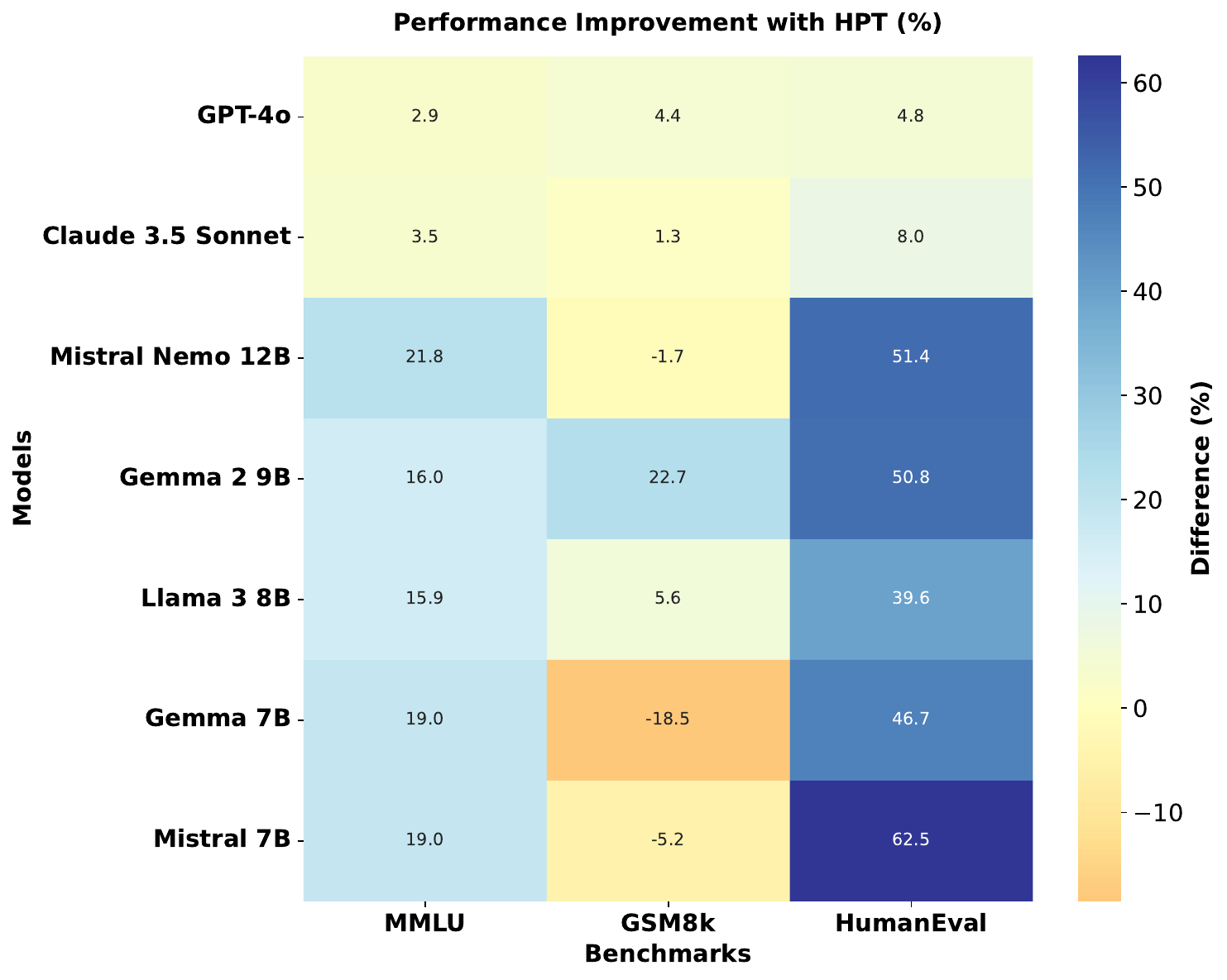}
    \caption{\textbf{Performance Comparison of HPT-based Evaluation vs. Standard Evaluation}: Performance improvements (in \%) when using HPT-based evaluation compared to standard evaluation across three benchmarks: MMLU, GSM8k, and HumanEval. Positive values indicate performance gains with HPT, while negative values indicate performance decreases. The baseline standard evaluation scores are sourced from Hugging Face leaderboard and official research reports.}
    \label{fig:3}
\end{figure}
\subsection{Results on Standard Benchmarks: MMLU, GSM8K, and Humaneval}
The evaluation of HPF effectiveness as shown in Figure \ref{fig:3} spans three standard benchmarks: MMLU, GSM8k, and HumanEval. On the MMLU benchmark, which tests general knowledge across multiple domains, all models showed notable improvements over their baseline performance. Mistral-Nemo 12B demonstrated the most substantial MMLU enhancement (+21.8\%), while Claude 3.5 Sonnet achieved a consistent improvement of 3.5\%. In mathematical reasoning, assessed through GSM8k, the results revealed a correlation with the model scale. Larger models like GPT-4 and Claude 3.5 Sonnet showed modest gains (+4.4\% and +1.3\% respectively), while smaller models exhibited more variable performance. The HumanEval benchmark, which assesses code generation capabilities, revealed the most dramatic improvements across all models. Mistral 7B achieved an exception 62.5\% improvement in HumanEval scores, followed by Mistral-Nemo 12B with an impressive 51.4\% improvement, and Gemma-2 9B with a 50.8\% enhancement. The results suggest that HPF enhances performance on all benchmarks for the majority of SLMs and achieves similar performance to LLMs such as GPT-4o and Claude 3.5 Sonnet, thereby addressing \textbf{RQ1}, its impact is particularly pronounced in programming tasks, suggesting that the technique may be especially valuable for enhancing code-related capabilities.
\begin{table*}[h]
\caption{HPI (lower is better) and accuracy of LLMs across MMLU, GSM8K, BoolQ, and CSQA datasets. \textcolor{blue}{Blue} indicates datasets where the LLM with the best HPI does not achieve the best performance. \textcolor{darkgreen}{Green} indicates the LLM with the best performance over the maximum number of datasets.}
    \centering
    \begin{tabular}{lcccccccc}
    \toprule
    \textbf{DATASETS} & \multicolumn{2}{c}{\textbf{MMLU}} & \multicolumn{2}{c}{\textbf{GSM8k}} & \multicolumn{2}{c}{\textbf{BoolQ}} & \multicolumn{2}{c}{\textbf{CSQA}} \\
    \cmidrule(r){2-3} \cmidrule(r){4-5} \cmidrule(r){6-7} \cmidrule(r){8-9}
    \textbf{Models}& \textbf{HPI} & \textbf{Accuracy} & \textbf{HPI} & \textbf{Accuracy} & \textbf{HPI} & \textbf{Accuracy} & \textbf{HPI} & \textbf{Accuracy} \\
    \midrule
    \textbf{GPT-4o} & \textcolor{blue}{1.81} & 91.61 & 1.71 & 96.43 & 1.32 & 96.82 & \textbf{1.65} & \textbf{92.54} \\
    \textcolor{darkgreen}{Claude 3.5 Sonnet} & 1.84 & \textcolor{blue}{{92.16}} & \textbf{1.35} & \textbf{97.72} & \textcolor{blue}{1.20} & 99.81 & 2.01 & 86.15   \\
    \textbf{Mistral-Nemo 12B} & 2.45 & 89.75  & 3.01 & 86.80 & 1.75 & \textcolor{blue}{99.87} & 2.06 & 90.17 \\
    \textbf{Gemma-2 9B} & 2.34 & 87.28 & 2.17 & 91.28 & 1.30 & 98.28 & 1.94 & 88.86 \\
    \textbf{Llama-3 8B} & 2.84  & 82.63 & 2.34 & 86.20 & 1.37 & 99.30 & 2.43 & 84.76 \\
    \textbf{Gemma 7B} & 2.93 & 83.31 & 6.70 & 27.88 & 1.45 & 99.42 & 2.50 & 83.78\\
    \textbf{Mistral 7B} &2.89 &81.45 & 5.11&46.93 & 1.41 & 98.07 & 2.49 & 82.06\\
    \bottomrule        
    \end{tabular}
    
    \label{tab:1}
\end{table*}

Table \ref{tab:1} highlights the improved performance of various LLMs on MMLU, with all models showing an HPI index below three. This indicates that reasoning over most MMLU samples requires minimal cognitive effort for these models, compared to baseline multi-shot CoT methods (5 shot), which typically require more than five examples and are more cognitively demanding according to HPT. Interestingly, while Claude 3.5 Sonnet achieves the highest MMLU accuracy, GPT-4o records the best HPI score, showing that minimal cognitive effort does not necessarily equate to the best performance addressing \textbf{RQ2}. The enhancement in GSM8k is relatively smaller compared to MMLU, with decreased performances for both Mistral 7B and Gemma 7B. The high HPI values for Gemma 7B and Mistral 7B indicate that none of the five prompting strategies in HPF posed significant cognitive challenges for these LLMs, i.e more cognitively demanding prompting strategies are needed, highlighting a limitation of the HPF. As shown in Table \ref{tab:2}, Claude 3.5 Sonnet achieves a perfect pass@1 of 1.00 with low HPI values, outperforming GPT-4o, which scores 0.95 but has a higher HPI. Gemma 7B struggles with the lowest pass@1 of 0.79 and the highest HPI of 3.71, indicating a need for a more complex prompting strategy.

Notably, HPF noticeably boosted the performance of the majority of LLMs on three benchmark datasets, despite the HPI difference being less than 1 compared to the top-performing LLMs. This suggests that even with a minimal number of inferences, utilizing HPF can achieve optimal performance, unlike multi-shot prompting and prompt optimization strategies, thereby addressing \textbf{RQ3}. This highlights that tailoring the prompting strategy to align with the complexity of each dataset instance can lead to substantial improvements, achieving performance levels comparable to state-of-the-art LLMs such as GPT-4o and Claude 3.5 Sonnet on these benchmarks.
\begin{table}[h]
    \caption{HPI (lower is better) and Pass@1 of LLMs on the HumanEval dataset. \textcolor{blue}{Blue} indicates datasets where the LLM with the best HPI does not achieve the best performance. \textcolor{darkgreen}{Green} indicates the LLMs with the best performance over the dataset.}
    \centering
    \begin{tabular}{lcccc}
    \toprule
    \textbf{DATASET} & \multicolumn{2}{c}{\textbf{HumanEval}} \\
    \cmidrule(r){2-3}
    \textbf{Models} & \textbf{HPI} & \textbf{Pass@1} \\
    \midrule
    \textbf{GPT-4o} & 2.25 & 0.95 \\
    \textcolor{darkgreen}{Claude 3.5 Sonnet} & 1.04 & \textcolor{blue}{1.00} \\
    \textbf{Mistral-Nemo 12B} & 2.07 & 0.96 \\
    \textbf{Gemma-2 9B} & \textcolor{blue}{1.01} & 0.91 \\
    \textcolor{darkgreen}{Llama-3 8B} & 1.03 & \textcolor{blue}{1.00} \\
    \textbf{Gemma 7B} & 3.71 & 0.79 \\
    \textbf{Mistral 7B} & 1.10 & 0.93 \\
    \bottomrule        
    \end{tabular}

    \label{tab:2}
\end{table}
\subsection{Results on Other Datasets}
Table \ref{tab:1} presents LLM performance on the BoolQ and CSQA datasets. While no significant insights emerge, an unexpected result is GPT-4o’s poor performance, which deviates from its typical trend. With most LLMs achieving near-perfect scores, BoolQ appears insufficiently complex to serve as an effective benchmark for modern LLMs, as they excel even with minimal cognitive prompting. This highlights HPF’s value in assessing dataset complexity relative to LLM capabilities, providing researchers with insights for designing more challenging and robust benchmarks.  
\begin{table*}[]
    \caption{HPI (lower is better), BLEU score for IWSLT, and ROUGE-L score for SamSum, of LLMs with thresholds.}
    \centering
    \begin{tabular}{p{3cm}p{1cm}p{1cm}p{1cm}p{1cm}p{1cm}p{1cm}p{1cm}p{1cm}}
    \toprule
    \textbf{DATASETS} & \multicolumn{4}{c}{\textbf{IWSLT}} & \multicolumn{4}{c}{\textbf{SamSum}} \\
    \cmidrule(r){2-5} \cmidrule(r){6-9}
     & \multicolumn{2}{c}{\textbf{HPI}} & \multicolumn{2}{c}{\textbf{BLEU}} & \multicolumn{2}{c}{\textbf{HPI}} & \multicolumn{2}{c}{\textbf{ROUGE-L}} \\
    \cmidrule(r){2-3} \cmidrule(r){4-5} \cmidrule(r){6-7} \cmidrule(r){8-9}
   \textbf{Models}  & \textbf{0.15} & \textbf{0.20} & \textbf{0.15} & \textbf{0.20} & \textbf{0.15} & \textbf{0.20} & \textbf{0.15} & \textbf{0.20} \\
    \midrule
    \textbf{GPT-4o} & \textbf{2.66} & \textbf{3.08} & \textbf{0.32} & \textbf{0.32} & \textbf{1.11} & \textbf{1.21} & \textbf{0.30} & \textbf{0.29} \\
    \textbf{Claude 3.5 Sonnet} & 4.63 & 4.87 & 0.20 & 0.20 & 1.25 & 1.60 & 0.23 & 0.23 \\
    \textbf{Mistral-Nemo 12B} & 2.87 & 3.40 & 0.27 & 0.27 & 1.19 & 1.47 & 0.23 & 0.24 \\
    \textbf{Gemma-2 9B} & 4.40 & 4.75 & 0.21 & 0.20 & 1.30 & 1.86 & 0.22 & 0.22 \\
    \textbf{Llama-3 8B} & 3.40 & 3.92 & 0.24 & 0.23 & 1.30 & 1.72 & 0.22 & 0.22 \\
    \textbf{Gemma 7B} & 5.39 & 5.84 & 0.08 & 0.09 & 3.31 & 5.03 & 0.11 & 0.10 \\
    \textbf{Mistral 7B} & 3.52 & 4.14 & 0.22 & 0.22 & 1.26 & 1.68 &0.21 & 0.22 \\
    \bottomrule        
    \end{tabular}

    \label{tab:3}
\end{table*}

Table \ref{tab:3} presents the performance of LLMs on IWSLT and SamSum datasets at varying thresholds. GPT-4o consistently achieved the highest scores across all thresholds, while most models, except Gemma 7B, performed similarly. Interestingly, Claude 3.5 Sonnet, which excelled in reasoning tasks, did not perform as strongly in summarization and translation tasks. The threshold selection is guided by the observed performance plateau across most LLMs as the threshold increases.
\subsection{Threshold Selection for SamSum and IWSLT}
\begin{figure}[hbt!]
    \centering
    \includegraphics[width=\linewidth]{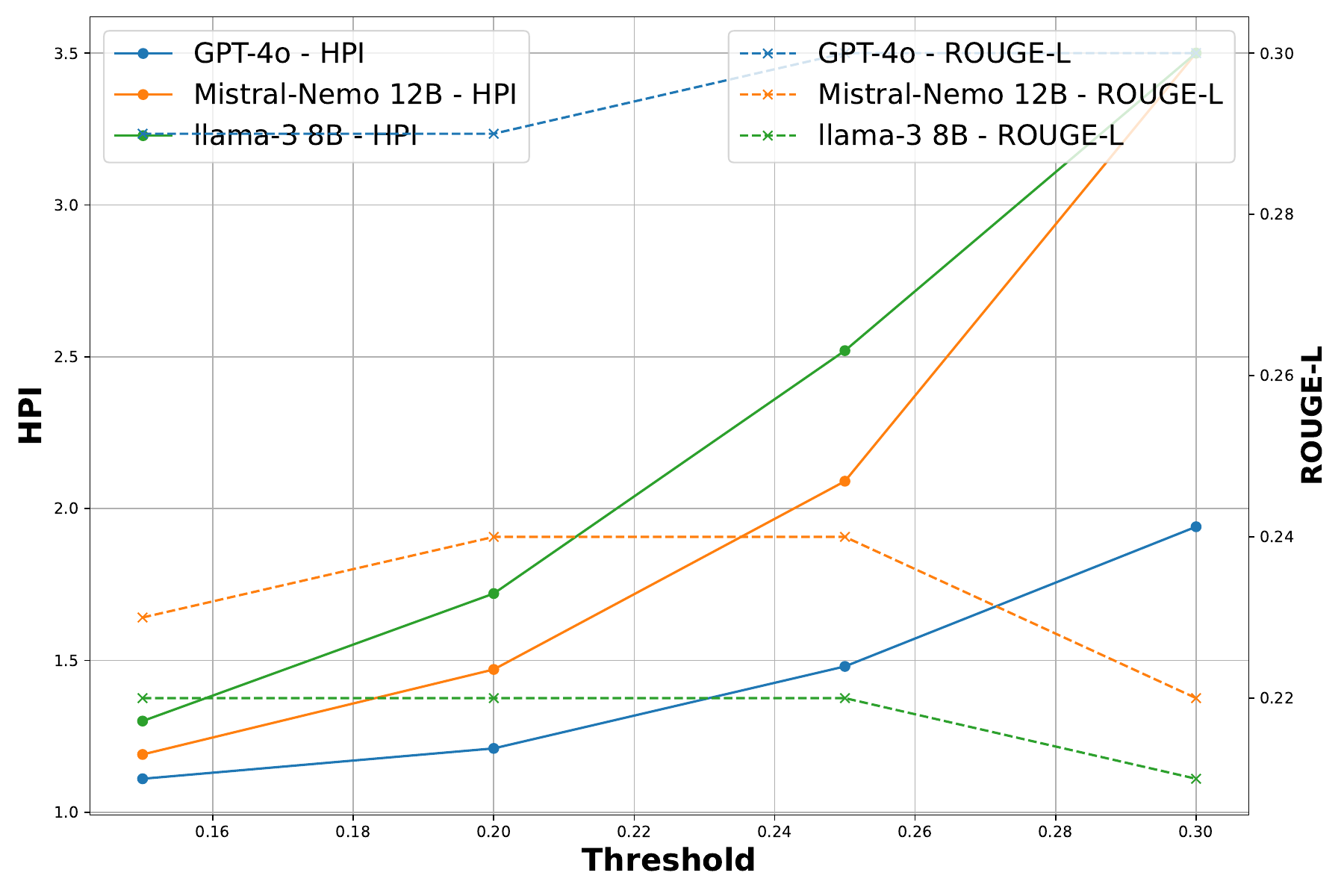}
    \caption{Comparison of HPI and ROUGE-L scores across different threshold values on SamSum dataset.}
    \label{fig:7}
\end{figure}

\begin{figure}[hbt!]
    \centering
    \includegraphics[width=\linewidth]{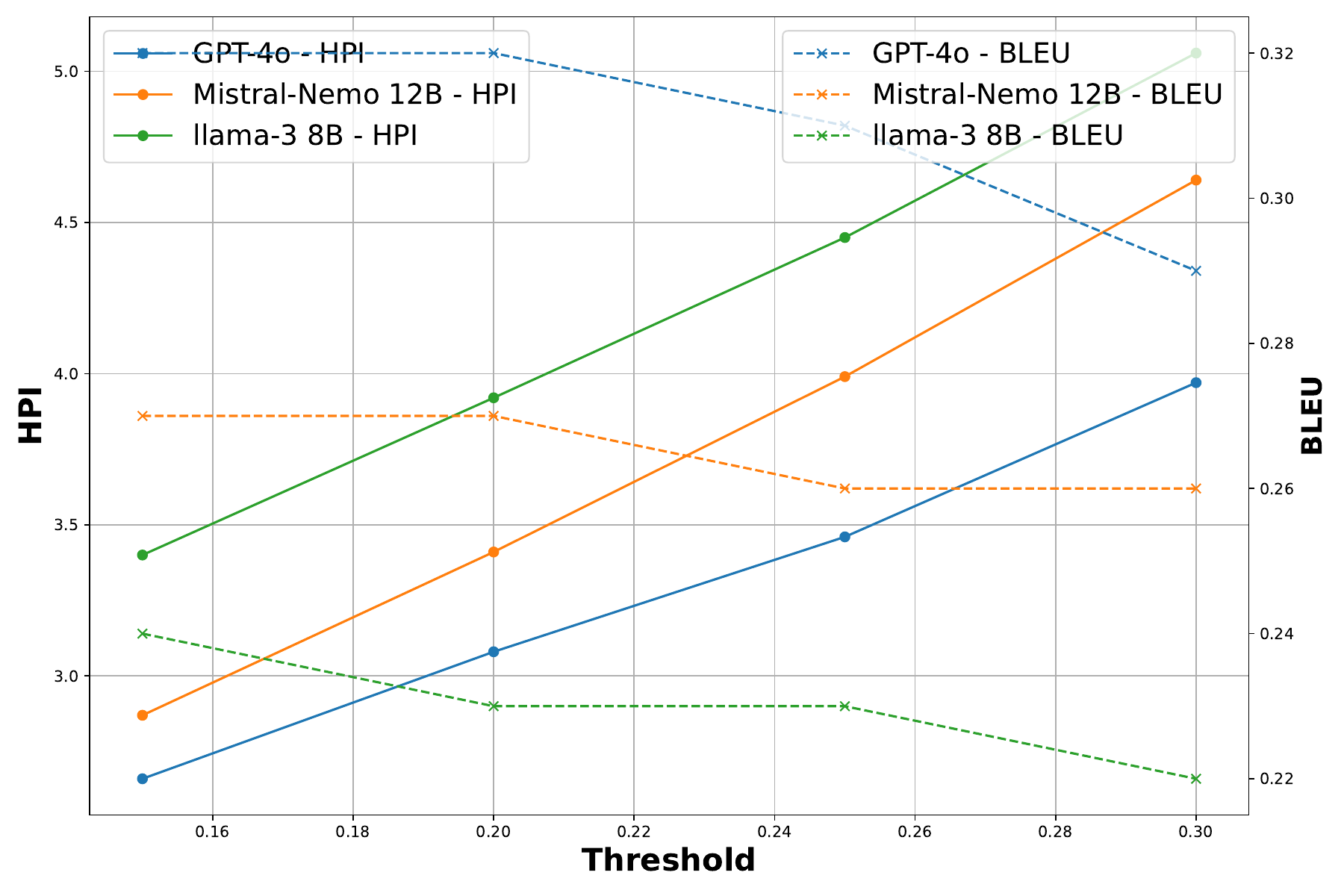}
    \caption{Comparison of HPI and BLEU score across different threshold values in the translation task.}
    \label{fig:8}
\end{figure}

In addition to the 0.15 and 0.20 thresholds presented in the main experiments, extended evaluations were conducted on the IWSLT and SamSum datasets using thresholds of 0.25 and 0.30 with GPT-4o, Mistral-Nemo 12B, and Llama-3 to assess the impact of varying thresholds on LLM performance.\\[5pt]
\textbf{SamSum Dataset: } In the summarization task, increasing the threshold evaluates an LLM’s ability to condense content while retaining key information. Higher thresholds like 0.25 and 0.30 reveal the trade-offs between conciseness and informativeness. However, as shown in Figure \ref{fig:7}, there was no significant improvement in ROUGE-L, except for a slight increase with GPT-4o. The experiments showed a sharp rise in HPI, reflecting the increased task complexity. These results suggest that LLM performance has plateaued, with no further gains at higher thresholds. This validates that the use of 0.15 and 0.20 thresholds are sufficient for optimal LLM performance.
\\[5pt]
\textbf{IWSLT Dataset: }In machine translation, higher thresholds (0.25 and 0.30) impose stricter evaluations, assessing how well models capture the nuances of the source text. Lower thresholds (0.15 and 0.20) focus on general adequacy, while higher ones test performance under more challenging conditions. As shown in Figure \ref{fig:8}, no BLEU improvements were observed across any LLMs, with models either reaching saturation or showing decreased performance alongside a rapid rise in HPI. This validates the selection of 0.15 and 0.20 thresholds are sufficient for optimal LLM performance.
\subsection{Complexity Levels with LLM-as-a-Judge}
\label{sec:llm-as-a-judge}
\begin{figure}[h]
    \centering
    \includegraphics[width =\columnwidth]{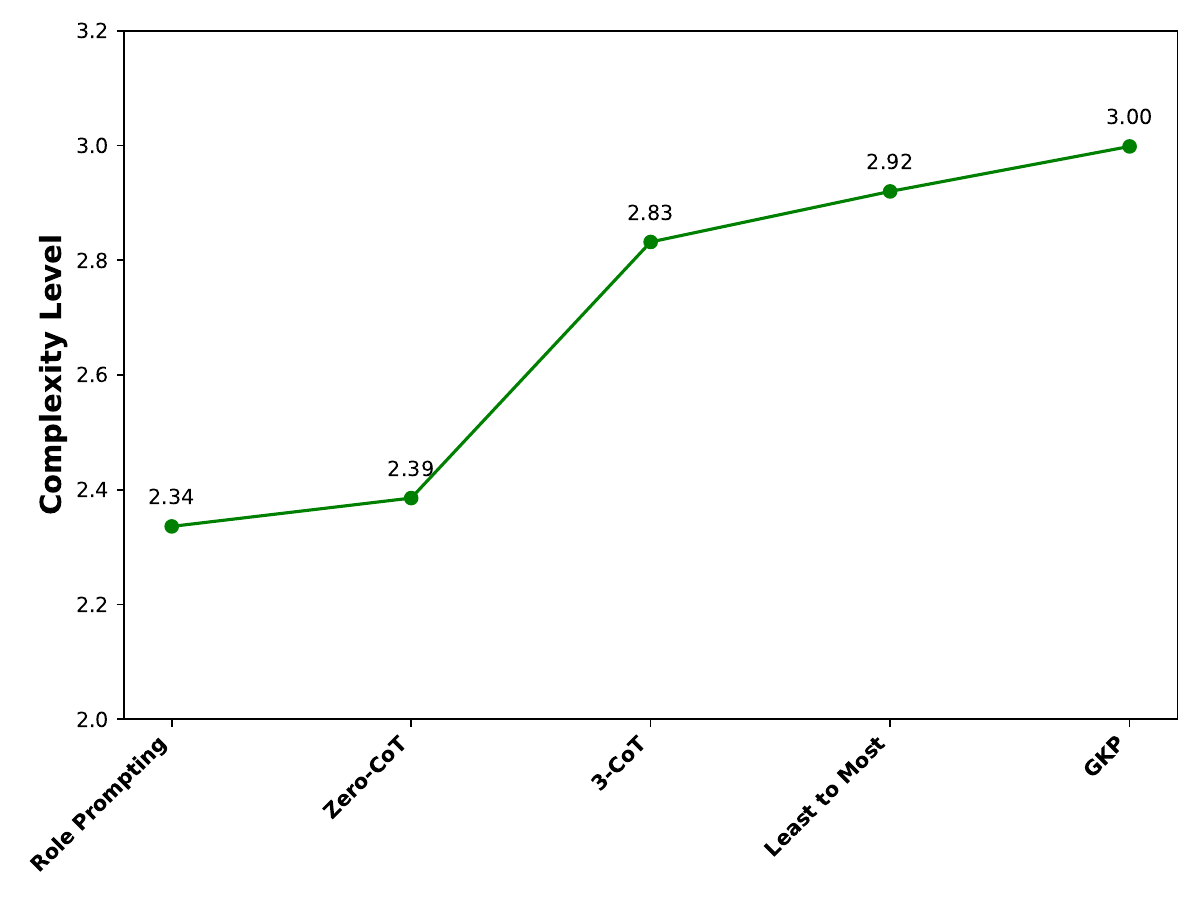}
    \caption{Hierarchy of prompting strategies with LLM-as-a-Judge framework with GPT-4o as the judge.}
    \label{fig:4}
\end{figure}

\begin{figure}
    \centering
    \includegraphics[width= \columnwidth]{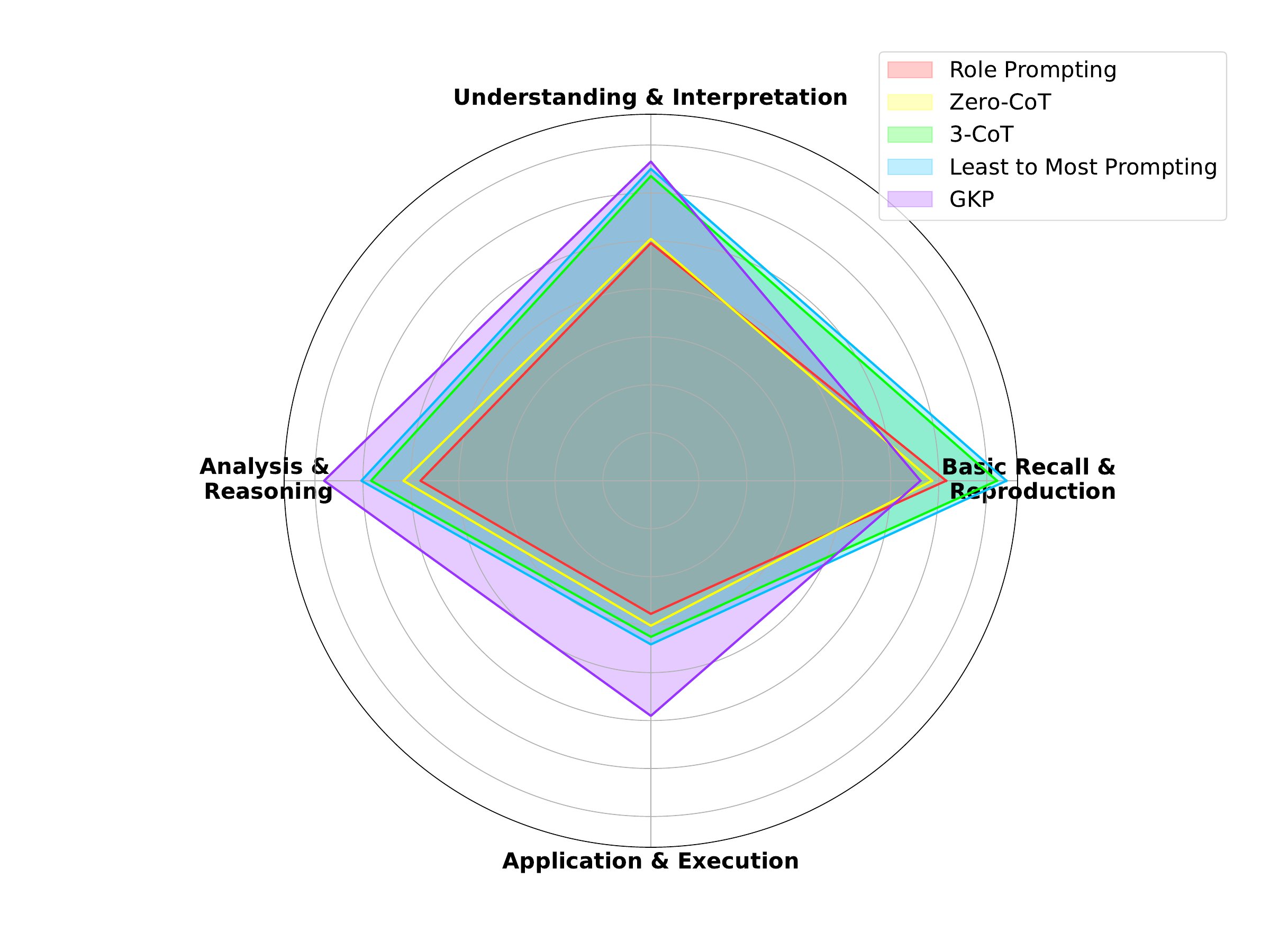}
    \caption{Scoring distribution for each of the four rules of the HPT for the prompting strategies in the HPF.}
    \label{fig:radar}
\end{figure}

This study evaluated prompting strategies by assessing how GPT-4o, as the LLM judge, replicates the hierarchical complexity levels of these strategies using a systematic scoring approach across tasks. Figure \ref{fig:4} shows a consistent hierarchy with less variability than human judges, indicating a strong alignment between LLM and human judgment. These results validate the proposed framework and demonstrate the correspondence between human cognitive principles and LLM behavior. Figure \ref{fig:radar} shows the scoring distribution across the four HPT rules for each strategy. Further details related to evaluation dataset specifications and scoring method are in Appendix \ref{appendix:llm-as-a-judge}.
\subsection{Parallels with System 1 and System 2 Thinking}
HPF parallels dual-process cognitive theories' System 1 and System 2 thinking \cite{booch2021thinking, kahneman2011thinking}. HPT classifies tasks, and HPF designs prompts based on cognitive complexity, reflecting human cognitive resource allocation. For tasks with low cognitive demands, HPF uses simple prompts akin to System 1 thinking, like fact recall or basic classification, enabling quick LLM responses with minimal reasoning. Conversely, tasks with high cognitive demands require prompts for complex reasoning and problem-solving, similar to System 2 thinking, involving logical arguments or intricate problems needing deliberate processing. Elevated HPF levels are used for tasks demanding deep analysis.

HPF explicitly measures this transition with HPI, assessing the cognitive load required for each task. By tailoring prompting strategies to task complexity, HPF optimizes LLM performance, much like humans adaptively switch between System 1 and System 2 based on the situation. This parallel highlights how HPT bridges computational strategies with human-like cognitive models, enabling more nuanced task evaluation and resource allocation.

\subsection{Adaptive HPF}
\begin{figure}[hbt!]
    \centering
    \includegraphics[width =\columnwidth]{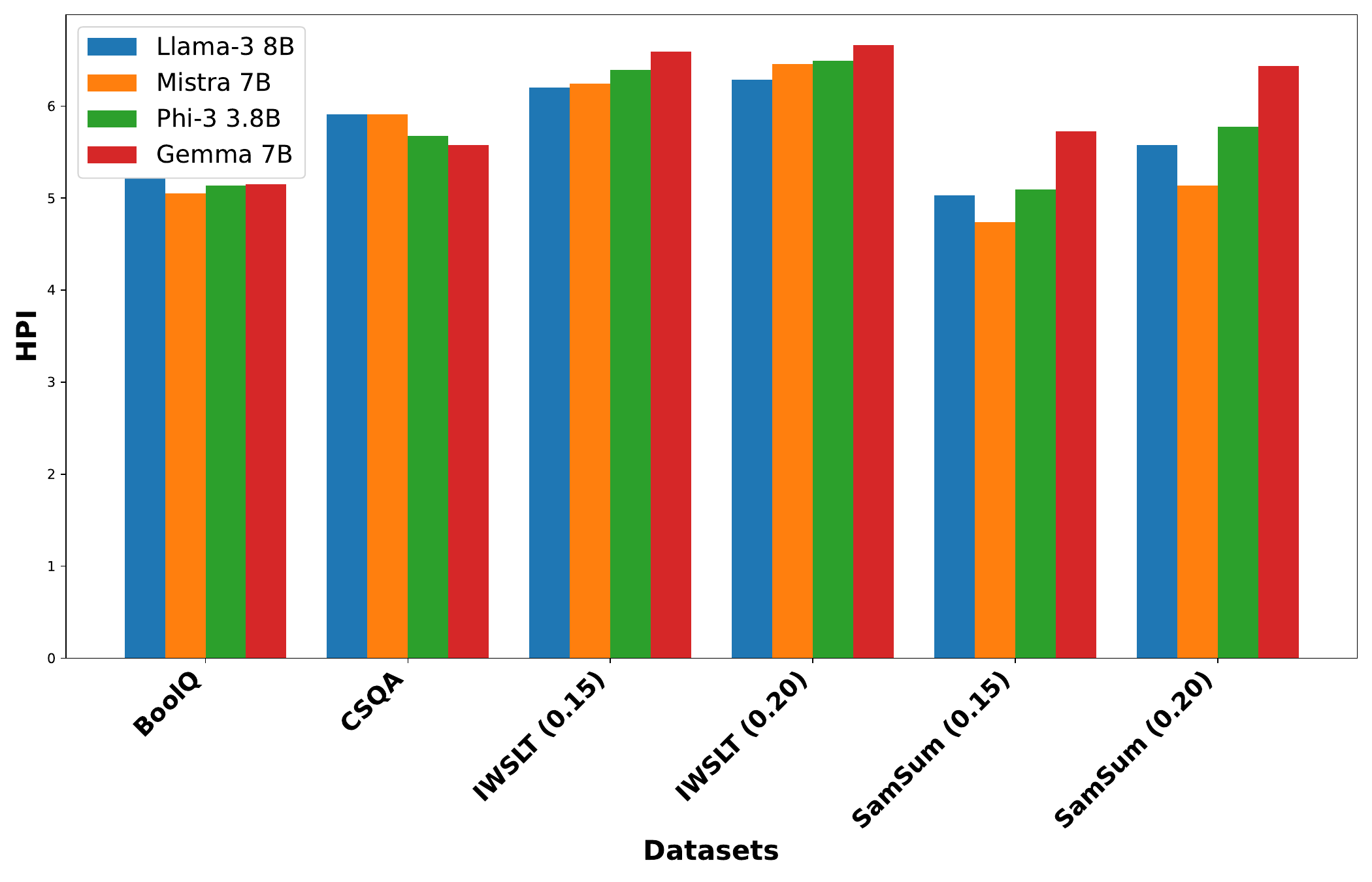}
    \caption{HPI of datasets for LLMs in Adaptive HPF.}
    \label{fig:6}
\end{figure}

The Adaptive HPF automates the selection of the optimal complexity level in the HPF using a \textit{prompt-selector}, Llama-3 8B in a zero-shot setting, bypassing iterative steps. Figure \ref{fig:6} shows that Adaptive HPF yields higher HPI but lower evaluation scores than the standard HPF. This suggests that Adaptive HPF struggles to select the optimal complexity level, likely due to hallucinations by the \textit{prompt-selector} when choosing the prompting strategy. For more results and ablation studies, see Appendix \ref{appendix: AHPF}.

The \textit{prompt-selector} can dynamically select the most suitable prompting strategy for a given task's complexity from the HPF's hierarchy of complexity levels. To determine the most effective prompting strategy to complete the task, the \textit{prompt-selector} was given a maximum number of iterations equivalent to the number of levels in the manual HPF. The score for \emph{i}th iteration is \emph{i + x}, where \emph{x} is the complexity level by the \textit{prompt-selector}. If the LLM fails to complete the task after all iterations, it is assigned a penalty, $\texttt{HPI}_{Dataset}$.
\begin{algorithm}
\caption{HPI Computation for Adaptive HPF}\label{alg:2}
\begin{algorithmic}
    \State $\texttt{HPI\_List} = []$  
    \For{sample $j$ in the evaluation dataset} 
        \State $\texttt{solved} = \texttt{False}$  
        \For{iteration $i = 1$ to $m$}  
            \State Select prompting strategy at level $x$
            \If{LLM completes the task at iteration $i$}
                \State $\texttt{HPI\_List}[j] = x + i$  
                \State $\texttt{solved} = \texttt{True}$  
                \State \textbf{break}  
            \EndIf
        \EndFor
        \If{\texttt{solved} = \texttt{False}}  
            \State $\texttt{HPI\_List}[j] = m + \texttt{HPI}_{\texttt{Dataset}}$
        \EndIf
    \EndFor
    \State  $\texttt{HPI}_{\texttt{Adaptive}} = \frac{1}{n} \sum_{j=1}^{n} \texttt{HPI\_List}[j]$
\end{algorithmic}
\end{algorithm}
Algorithm \ref{alg:2} demonstrates the calculation of HPI for an adaptive HPF, where $x$ denotes the HPF level chosen by the \textit{prompt-selector} at the \(i\)th iteration as the task is being tackled. Here, \(m\) indicates the total number of HPF levels, and \(n\) signifies the total quantity of samples in the evaluation set.
\section{Conclusion}  
\label{sec:Conclusion}  
The HPT offers an efficient way to evaluate LLMs by focusing on task cognitive demands. It shows that cognitively inspired selection of prompting strategies enhances LLM performance across various datasets. This method offers insights into LLM problem-solving and improves evaluation methods based on human cognition, supporting better in-context learning strategies for assessing LLMs.

\section{Limitations}
\label{sec:Limitations}
\textbf{Human Annotation Constraints}: A limitation of this study is the reliance on human evaluation for inducing the $\texttt{HPI}_{Dataset}$ penalty into the HPF. While this study assessed 5\% of the datasets, expanding coverage would offer a more comprehensive analysis. However, due to constraints in human resources for manual annotation, we could not include a larger portion. Future work could address this by increasing manpower or automating parts of the evaluation process.\\[5pt]
\textbf{HPF Optimization}: The effectiveness of the HPF heavily relies on the quality of the prompts used at each level of the taxonomy. Crafting high-quality prompts that accurately reflect the subtleties of each level demands considerable expertise and repeated refinement. This study only investigated a limited set of prompting strategies within the HPF, indicating a need for further research into creating diverse structural frameworks and incorporating additional prompting strategies.\\[5pt]
\textbf{Zero-shot Prompt Selection}: HPF dynamically determines the optimal cognitive complexity level by iterating through the framework’s levels, which leads to increased inference time. While this study investigated Adaptive HPF for zero-shot prompt selection, it faced considerable hallucinations. Future research should focus on automating HPF using fine-tuning or reinforcement learning-based approaches to select the appropriate complexity level without manual iteration. This strategy would optimize inference time and improve overall performance.
\section{Ethical Statement}
\label{sec:ethical impact}
The $\texttt{HPI}_{Dataset}$ assigned by experts to MMLU, GSM8k, HumanEval, BoolQ, CSQA, IWSLT, and SamSum may introduce bias due to the subjective nature of expert scoring, influenced by individual experience and perspective. However, these publicly available, widely recognized datasets help mitigate unforeseen ethical concerns. Acknowledging potential scoring bias remains essential for transparency and integrity in the analysis. 
\bibliographystyle{ACM-Reference-Format}
\bibliography{sample-base}

\appendix

\section{Human Annotation and Judgement Policy}
\label{appendix: human policy}
\subsection{Human Annotation Policy}
$\texttt{HPI}_{Dataset}$ is introduced to penalize the HPI of tasks or samples unsolvable by the LLM, aligning the framework more closely with human cognitive demands and enhancing its comprehensiveness. We implemented a rigorous human annotation process to ensure the quality of $\texttt{HPI}_{Dataset}$ scored by human experts for the datasets. Human annotators were tasked with calculating the HPI for each sample in a given dataset. The HPI quantifies the cognitive demands imposed on human expert proficiency in completing a task, based on the HPT, where higher values indicate greater cognitive demands. Each sample was scored on a scale from \emph{1} (lowest complexity level) to \emph{5} (highest complexity level) for the following criteria:
\begin{enumerate}
    \item \textbf{Basic Understanding and Reproduction}: This criterion evaluates the annotator's ability to comprehend and accurately reproduce the content.
    
    \item \textbf{Understanding and Interpretation}: This criterion assesses the annotator's depth of understanding and the ability to interpret the information correctly.
 
    \item \textbf{Analysis and Reasoning}: This criterion measures the annotator's ability to analyze the information and apply logical reasoning.

    \item \textbf{Application of Knowledge and Execution}: This criterion evaluates the annotator's practical application of knowledge and the execution of tasks based on the relevant knowledge.
\end{enumerate}
Higher scores for the four rules signify a stronger influence of the respective rules, indicating that completing the task requires greater cognitive effort. The $\texttt{HPI}_{Dataset}$ for each dataset, as shown in Table \ref{tab:4}, was calculated by taking the mean of the values from these four criteria, acknowledging the challenge of estimating or computing the individual weights of the influence of each rule. \\[5pt]
The \texttt{Representative Set Size} in Table \ref{tab:4} refers to the subset of the dataset evaluated by human annotators, ensuring that the assessment reflects the overall task. Human annotation, while time-consuming and costly, provides a gold standard for calibrating the evaluation process of this paper. Selecting 5\% of the dataset as the representative set size balances quality assessment and feasibility, capturing the dataset’s diversity and ensuring that human annotations encompass a broad range of cases without needing to annotate every sample.
\begin{table}[h]
\caption{\textbf{$\texttt{HPI}_{Dataset}$ scores across datasets evaluated by human annotators}. The table lists the evaluation set size, representative set size, and $\texttt{HPI}_{Dataset}$ for various datasets. $\texttt{HPI}_{Dataset}$ scores provide a measure of task complexity relative to human annotators.}
    \centering
          
    \begin{tabular}{p{1.7cm}p{1.5cm}p{2.2cm}p{1.3cm}}
    \toprule
    \textbf{Dataset} & \textbf{Evaluation Set Size} & \textbf{Representative Set Size} & \textbf{$\texttt{HPI}_{Dataset}$} \\
    \midrule
    \textbf{MMLU}    & 14500 & 725  &3.03 \\
    \textbf{GSM8k}   & 1320 & 66 & 2.14\\
    \textbf{Humaneval} & 160 & 8 & 4.68 \\
    \textbf{BoolQ} & 3270 & 162 & 1.71 \\
    \textbf{CSQA} & 1221 & 60 & 2.52 \\
    \textbf{IWSLT} & 890 & 45 & 1.92 \\
    \textbf{SamSum} & 819 & 40 & 2.23 \\
    \bottomrule
    \end{tabular}

    \label{tab:4}
\end{table}
\subsection{Human Judgement Policy}
To populate the HPF with relevant prompting strategies across a wide range of strategies, human annotators who adhered to the annotation policy for assessing $\texttt{HPI}_{Dataset}$ were instructed to follow a judgment policy for a predefined set of prompting strategies. They were instructed to evaluate the influence of the four rules of the HPT on solving the annotated tasks using each prompting strategy, rating their influence as High (H), Moderate (M), or Low (L). It’s important to note that a high rating on rule \ref{4} has a greater influence than a high rating on rule \ref{3}, and similarly for the other two rules. Considering the rating as shown in Table \ref{tab:5} and varying influences of these rules, five prompting strategies that prioritize comprehensive coverage of cognitive demands while ensuring the set optimally widens the variation across complexity levels were selected for populating the HPF.
\begin{table}[h]
    \centering  
    \begin{tabular}{p{2.3cm}p{1cm}p{1cm}p{1cm}p{1cm}}
    \toprule
    \textbf{Prompting Strategy} & \textbf{Rule 1} & \textbf{Rule 2} & \textbf{Rule 3} & \textbf{Rule 4} \\
    \midrule
    \textcolor{darkgreen}{Role Prompting} & L & L & L & L\\
    \textbf{Emotion Prompting} & L & L & M & L\\
    \textcolor{darkgreen}{Zero-shot CoT} & L & L & M & L\\
    \textbf{Meta Prompting} & M & H & M & L\\
    \textcolor{darkgreen}{Three-shot CoT} & H & H & M & L\\
    \textbf{Five-shot CoT} & H & H & H & L\\
    \textbf{Chain-of-Verification} & H & H & H & H\\
    \textcolor{darkgreen}{Least-to-Most Prompting} & H & H & H & L\\
    \textbf{Self-Consistency} & H & H & H & M\\
    \textcolor{darkgreen}{GKP} & L & H & H & H\\
    \bottomrule
    \end{tabular}
\caption{Human judgment of influence of the rules of taxonomy on different prompting strategies in solving the tasks of the representative set. The ratings are provided based on a voting system involving all human annotators. \textcolor{darkgreen}{Green} represents the prompting strategies selected for populating the complexity levels of the HPF.}
    \label{tab:5}
\end{table}
\section{LLM-as-a-Judge}
\label{appendix:llm-as-a-judge}
\subsection{Scoring Prompt Template}
The system prompt is designed to guide the LLM judge in evaluating different prompting strategies based on four specific criteria: Basic Recall and Reproduction, Understanding and Interpretation, Analysis and Reasoning, and Application of Knowledge and Execution. Each criterion is scored on a scale of 1-5. The evaluation uses GPT-4o as a judge, with the following system prompt:\\[5pt]
\texttt{You are a judge evaluating different prompting strategies and you need to score these prompting strategies based on pre-defined criteria. Different prompting strategies leverage varied amounts of intelligence from the model to achieve the required answer. So, assign the scores very carefully based on your analysis of the prompt and its effect on your intelligence to achieve the given answer as well as the number of multi-step prompts which increases the complexity of execution.\\[5pt]
score1: Basic Recall and Reproduction\\
Definition: The need of the model to remember and reproduce factual information without interpretation or analysis to answer the prompt\\
Range: 1-5\\[5pt]
score2: Understanding and Interpretation\\
Definition: The need of the model to comprehend and explain the meaning of information, summarizing or clarifying content to answer the prompt\\
Range: 1-5\\[5pt]
score3: Analysis and Reasoning\\
Definition: The need for the model to break down complex information, understand relationships, and solve problems using logical reasoning to answer the prompt\\
Range: 1-5\\[5pt]
score4: Application of Knowledge and Execution\\
Definition: The need for the model to apply knowledge in practical situations, execute multi-step processes, and solve complex tasks to answer the prompt\\
Range: 1-5}
\subsection{Hybrid Dataset}
The hybrid dataset is composed of 1106 samples uniformly distributed over seven different task-specific datasets, covering a wide range of language understanding and generation tasks. This diversity allows for a comprehensive evaluation of the prompting strategies across various problem types. The evaluation uses a hybrid dataset composed of samples from various task-specific datasets and each dataset contributes specific types of tasks:
\begin{enumerate}
    \item MMLU (Massive Multitask Language Understanding)
    \item HumanEval (Code Generation and Completion)
    \item GSM8K (Grade School Math 8K)
    \item BoolQ (Boolean Questions)
    \item CSQA (Commonsense Question Answering)
    \item IWSLT (International Workshop on Spoken Language Translation)
    \item SamSum (Dialogue Summarization)
\end{enumerate}
\subsection{Scoring Method}
For each prompting strategy (Role Prompting, Zero-shot CoT, Three-shot CoT, Least to Most Prompting, Generated Knowledge Prompting), the system:
\begin{enumerate}
    \item Applies the prompting strategy to each sample in the hybrid dataset
    \item Generates an answer using GPT-4o
    \item Presents the prompt, generated answer, and correct answer to the LLM judge
    \item Collects scores for each of the four criteria and the system calculates average scores for each criterion across all tasks and datasets.
\end{enumerate}
This study ensured that both the human judge and the LLM judge utilized the same scoring methodology to eliminate any potential bias in the comparison.
\begin{table*}[hbt!]
  \caption{HPI (lower is better) of LLMs across datasets (with thresholds) for Adaptive HPF.}
    \centering  
    \begin{tabular}{lcccccc}
        \toprule
        \textbf{Model} & \textbf{BoolQ} & \textbf{CSQA} & \textbf{IWSLT (0.15)} & \textbf{IWSLT (0.20)} & \textbf{SamSum (0.15)} & \textbf{SamSum (0.20)} \\
        \toprule
        \textbf{Llama-3 8B} & 5.2173 & 5.9136 & \textbf{6.2006} & \textbf{6.2841} & 5.0316 & 5.5756 \\
        \textbf{Mistra 7B} & \textbf{5.0483} & 5.9073 & 6.2478 & 6.4604 & \textbf{4.7423} & \textbf{5.1336} \\
        \textbf{Phi-3 3.8B} & 5.1386 & \textbf{5.6793} & 6.3955 & 6.4936 & 5.0961 & 5.7778 \\
        \textbf{Gemma 7B} & 5.1514 & 5.5771 & 6.5947 & 6.6605 & 5.7229 & 6.4347 \\
        \bottomrule
    \end{tabular}
  
    \label{tab:6}
\end{table*}
\begin{table*}[ht!]
\caption{\label{tab:7}Performance scores of LLMs across datasetsfor Adaptive HPF.}
  \centering   
  \begin{tabular}{p{2.75cm} p{1.75cm} p{1.5cm}p{1.75cm} p{1.75cm} p{1.75cm} p{2cm}}
    \toprule
    \textbf{Dataset} & \textbf{Metric} & \textbf{Threshold} & \textbf{Llama-3 8B} & \textbf{Phi-3 3.8B} & \textbf{Mistral 7B} & \textbf{Gemma 7B} \\ 
    \toprule
    \textbf{BoolQ} & Accuracy & - & 0.88577 & 0.91115 & \textbf{0.91752} & 0.91166 \\ 
    \textbf{CSQA} & Accuracy & - & 0.59451 & 0.68019 & 0.60111 & \textbf{0.68549} \\ 
    \multirow{2}{*}{\textbf{IWSLT}} & \multirow{2}{*}{BLEU} & 0.15 & \textbf{0.21140} & 0.15557 & 0.20000 & 0.08447 \\  
     & & 0.2 & \textbf{0.21146} & 0.15354 & 0.20568 & 0.07730 \\ 
    \multirow{2}{*}{\textbf{SamSum}} & \multirow{2}{*}{ROUGE-1} & 0.15 & 0.24407 & 0.20586 & \textbf{0.26910} & 0.16023 \\  
     & & 0.2 & 0.24981 & 0.21580 & \textbf{0.28335} & 0.16179\\
    \bottomrule
  \end{tabular}
  
\end{table*}
\section{Hallucination in Adaptive HPF}
\label{appendix: AHPF}
Hallucinations in \textit{prompt-selector} refer to instances where the LLM generates incorrect or misleading prompting levels or nonsensical information that is not supported by the HPF. These hallucinations can occur across various tasks, including question answering, multiple-choice questions, translation, and summarization. \\[5pt] For the BoolQ task, the \textit{prompt-selector} initially struggles, indicated by the iterations where it reaches Level 4 with hallucinations. However, by the fourth iteration, Llama-3 8B manages to answer correctly at Level 2. For the CSQA task, \textit{prompt-selector} exhibits hallucinations initially, shown by Level 4 and Level 0 (not included in HPF) responses. Eventually, it corrects itself by the third iteration, providing the correct answer at Level 2. For the IWSLT task, \textit{prompt-selector} demonstrates a consistent pattern of hallucinations across multiple iterations. Even though Llama-3 8B attempts the translation at Level 2 multiple times, it ultimately fails to provide a correct translation, indicating a persistent hallucination. For the SamSum task, \textit{prompt-selector} shows initial hallucinations in the first three iterations (Level 4). However, by the fourth and fifth iterations, the \textit{prompt-selector} starts producing lower levels. Finally, Llama-3 8B achieves the correct answer at Level 2 in the last iteration .\\[5pt]
The results in Table \ref{tab:6} and Table \ref{tab:7} indicate that the \textit{prompt-selector} exhibits hallucinations in selecting complexity levels across various tasks and iterations resulting in higher HPI for Adaptive HPF, with performance varying significantly. While the LLM can eventually produce correct answers, as seen in the BoolQ and SamSum tasks, it often requires multiple attempts and may still fail in tasks like IWSLT translation. 
\subsection{Prompt Template for Prompt-Selector}
The \textit{prompt-selector} in adaptive HPF selects the prompting level based on the task complexity to address the task. Llama-3 8B serves as the \textit{prompt-selector} in the experiments. The prompt template was meticulously designed to ensure maximum clarity, aiming to reduce hallucinations and select the most effective prompting strategy.\\[5pt]
\textbf{Prompt Template}: Choose the most effective prompting strategy among five available strategies for the task. Begin with the lowest indexed strategy and progress to higher indexed strategies if the earlier ones are not effective. For a given task, the prompting strategies are:
\begin{itemize}
    \item \textbf{Role Prompting:} Defines a role for the model in solving the task.
    \item \textbf{Zero-shot Chain of Thought prompting:} Stimulates reasoning and problem-solving by including the phrase 'Let's think step by step' without offering previous examples related to the task.
    \item \textbf{Three-shot Chain of Thought prompting:} Offers three examples related to the task to guide the model's reasoning process.
    \item \textbf{Least-to-most prompting:} Uses a sequential method to derive essential insights from the task to solve it.
    \item \textbf{Generated Knowledge Prompting:} Integration and application of external knowledge to accomplish the task. The external knowledge is generated using some other model based on the task.
\end{itemize}
Select only the index (do not provide the name) of the most effective prompting strategy.
\section{Computational Budget}
\label{appendix: budget}
All evaluation experiments and ablation studies were conducted on V100 GPUs (16GB and 32GB variants), utilizing a total of around 9,000 computation hours, this equates to approximately 1.125 petaflop-hours of computational resources.
\section{Large Language Models Used for Evaluation}
\label{appendix: LLMs}
The HPF supports leading open source and proprietary LLMs and includes mechanisms for optimizing performance through advanced quantization techniques. The experiments were conducted on the following instruction-tuned LLMs, and the model description and licenses are discussed in Table \ref{table:licenses}.\\[5pt]
\begin{table*}[!ht]
\caption{License information for LLMs used in the experiments.}
\centering  
\begin{tabular}{p{3cm}p{3.75cm}p{8cm}}
\toprule
\textbf{Model}       & \textbf{License Type}         & \textbf{Usage Restrictions} \\
\midrule
GPT-4o           &       Proprietary      &  Commercial use requires paid API access, subject to OpenAI's terms of service\\
Claude 3.5 Sonnet          &     Proprietary    &  Commercial use requires paid API access, subject to Anthropic's terms of service\\
Mistral-Nemo 12B          &        Proprietary            &Usage likely restricted to authorized partners or specific use cases \\
Gemma-2 9B             &   Research License     & Non-commercial use only, research purposes \\
Llama-3 8B             &     Research License    & Specific restrictions may apply, typically for non-commercial research use\\
Mistral 7B           & Open-source                   & Broad use allowed, must include original license and notices \\
Gemma 7B             & Open-source/Research       & Depending on the license, may have non-commercial restrictions or broad use allowed \\
Phi-3 3.8B           & Open-source         & Broad use allowed, must include original license and notices \\
\bottomrule
\end{tabular}

\label{table:licenses}
\end{table*}
The LLMs were loaded in 4-bit precision format, with a maximum generation limit of 1024 tokens per run to ensure concise outputs. The temperature was set to 0.6 to control prediction randomness, while top-p sampling (p=0.9) enabled the exploration of diverse continuations. Additionally, a repetition penalty was applied to discourage the generation of repeated phrases, promoting coherent and varied text output.
\section{Prompt Templates}
\subsection{Level 1: Role Prompting}
Role prompting represents the most basic interaction with an LLM, assigning it a specific role or task without additional context or examples. This approach relies solely on the initial instruction to guide responses. For instance, asking the LLM to \textit{“act as a translator”} prompts it to translate text based on its training data. While straightforward, this method may lack depth, resulting in less accurate or nuanced outputs. Table \ref{tab:prompt_role} shows the prompt templates used for role prompting across all datasets in the experiments.
\begin{table*}[!ht]
\caption{Prompt templates of different datasets for Role Prompting.}
    \centering
    \begin{tabular}{p{3cm}p{12cm}}
        \toprule
        \textbf{Dataset} & \textbf{Prompt} \\
        \midrule
        BoolQ & \texttt{Based on the passage: ``\textbf{passage}", answer True/False to the question: ``\textbf{question}" as an Omniscient person.} \\
        \midrule
        CSQA & \texttt{Choose the answer: ``\textbf{question}",A. ``\textbf{option 1}",B. ``\textbf{option 2}",C. ``\textbf{option 3}",D. ``\textbf{option 4}",E. ``\textbf{option 5}" as a critical thinker.} \\
        \midrule
        IWSLT & \texttt{Translate ``\textbf{english text}" to french as a Translator.} \\
        \midrule
        SamSum & \texttt{Summarize the Dialogue: ``\textbf{dialogue}" as a Summarizer.} \\
        \midrule
        GSM8k & \texttt{Based on the question: "\textbf{question}", calculate the numerical answer to the question as an expert mathematician.}\\
        \midrule
        HumanEval & \texttt{Complete the given code based on the mentioned constraints: "\textbf{code}" as an expert programmer.}\\
        \midrule
        MMLU & \texttt{Choose the answer: ``\textbf{question}",A. ``\textbf{option 1}",B. ``\textbf{option 2}",C. ``\textbf{option 3}",D. ``\textbf{option 4}" as a critical thinker.}\\
        \bottomrule
    \end{tabular}
    
    \label{tab:prompt_role}
\end{table*}
\subsection{Level 2: Zero-shot Chain-of-Thought Prompting}
Zero-shot Chain-of-Thought (CoT) prompting enhances basic role prompting by requiring the LLM to generate a reasoning process for a task, despite not being explicitly trained on similar examples. This method encourages the LLM to break down the problem and solve it step-by-step using its internal knowledge, improving response quality through logical progression and coherence. Table \ref{tab:prompt_zcot} displays the prompt templates used for Zero-CoT across all datasets in the experiments.
\begin{table*}[!ht]
\caption{Prompt templates of different datasets for Zero-shot Chain-of-Thought Prompting.}
    \centering
    \begin{tabular}{p{3cm} p{12cm}}
        \toprule
        \textbf{Dataset} & \textbf{Prompt} \\
        \midrule
        BoolQ & \texttt{Based on the passage: ``\textbf{passage}", answer True/False to the question: ``\textbf{question}". Let's think step by step.} \\
        \midrule
        CSQA & \texttt{Choose the answer: A. ``\textbf{option 1}",B. ``\textbf{option 2}",C. ``\textbf{option 3}",D. ``\textbf{option 4}",E. ``\textbf{option 5}". Let's think step by step.} \\
        \midrule
        IWSLT & \texttt{Translate ``\textbf{english text}" to french. Let's think step by step.} \\
        \midrule
        SamSum & \texttt{Summarize the Dialogue: ``\textbf{dialogue}". Let's think step by step.} \\
        GSM8k & \texttt{Based on the question: "\textbf{question}", calculate the numerical answer to the question. Let's think step by step.}\\
        \midrule
        HumanEval & \texttt{Complete the given code based on the mentioned constraints: "\textbf{code}". Let's think step by step.}\\
        \midrule
        MMLU & \texttt{Choose the answer: ``\textbf{question}",A. ``\textbf{option 1}",B. ``\textbf{option 2}",C. ``\textbf{option 3}",D. ``\textbf{option 4}". Let's think step by step.}\\
        \bottomrule
    \end{tabular}
    
    \label{tab:prompt_zcot}
\end{table*}
\subsection{Level 3: Three-Shot Chain-of-Thought Prompting}
Three-shot Chain-of-Thought (CoT) prompting builds on the zero-shot approach by providing the LLM with three task examples, including the reasoning steps used to reach the solution. These examples help the LLM grasp the required structure and logic, enabling it to better replicate the problem-solving process and produce more accurate, contextually relevant responses. Table \ref{tab:prompt_3cot} shows the prompt templates used for 3-CoT across all datasets in the experiments.
\begin{table*}[!ht]
\caption{Prompt templates of different datasets for Three-Shot Chain-of-Thought Prompting.}
    \centering   
    \begin{tabular}{p{2cm} p{12cm}}
        \toprule
        \textbf{Dataset} & \textbf{Prompt} \\
        \midrule
        \textbf{BoolQ} & \texttt{Based on the passage: "\textbf{passage1}", answer True/False to the question: "\textbf{question1}". Answer: "\textbf{answer1}". Explanation: "\textbf{explaination1}".
                        Based on the passage: "\textbf{passage2}", Answer True/False to the question: "\textbf{question2}". Answer: "\textbf{answer2}". Explanation: "\textbf{explaination2}". 
                        Based on the passage: "\textbf{passage3}", Answer True/False to the question: "\textbf{question3}". Answer: "\textbf{answer3}". Explanation: "\textbf{explaination3}". 
                        Based on the passage: "\textbf{passage}", Answer True/False to the question: "\textbf{question}".} \\
        \midrule
        \textbf{CSQA} & \texttt{Choose the answer: "\textbf{question1}",A. "\textbf{option1-1}",B. "\textbf{option2-1}",C. "\textbf{option3-1}",D. "\textbf{option4-1}",E. "\textbf{option5-1}", Answer: "\textbf{answer1}", Explanation: "\textbf{explaination1}".
                               Choose the answer: "\textbf{question2}", A. "\textbf{option1-2}",B. "\textbf{option2-2}",C. "\textbf{option3-2}",D. "\textbf{option4-2}",E. "\textbf{option5-2}", Answer: "\textbf{answer2}", Explanation: "\textbf{explainatio n2}". 
                               Choose the answer: "\textbf{question3}", A. "\textbf{option1-3}",B. "\textbf{option2-3}",C. "\textbf{option3-3}",D. "\textbf{option4-3}",E. "\textbf{option5-3}", Answer: "\textbf{answer3}", Explanation: "\textbf{explaination3}".
                               Choose the answer: "\textbf{question}", ``\textbf{question}",A. ``\textbf{option 1}",B. ``\textbf{option 2}",C. ``\textbf{option 3}",D. ``\textbf{option 4}",E. ``\textbf{option 5}".} \\
        \midrule
        \textbf{IWSLT} & \texttt{Translate "\textbf{english text1}" to French. French: "\textbf{french text1}". 
                                    Translate "\textbf{english text2}" to French. French: "\textbf{french text2}". 
                                    Translate "\textbf{english text3}" to French. French: "\textbf{french text3}". 
                                    Translate "\textbf{english text}" to French.} \\
        \midrule
        \textbf{SamSum} & \texttt{Summarize the Dialogue: "\textbf{dialogue1}". Summary: "\textbf{summary1}". 
                         Summarize the Dialogue: "\textbf{dialogue2}". Summary: "\textbf{summary2}". 
                         Summarize the Dialogue: "\textbf{dialogue3}". Summary: "\textbf{summary3}". 
                         Summarize the Dialogue: "\textbf{dialogue}".} \\
        \midrule
        \textbf{GSM8k} & \texttt{Based on the question:"\textbf{gsm8k\_question1}", calculate the numerical answer to the question.Answer: "\textbf{gsm8k\_ans1}".Based on the question:"\textbf{gsm8k\_question2}", calculate the numerical answer to the question.Answer: "\textbf{gsm8k\_ans2}". Based on the question:"\textbf{gsm8k\_question3}", calculate the numerical answer to the question.Answer: "\textbf{gsm8k\_ans3}". Based on the question: "\textbf{question}", calculate the numerical answer to the question.}\\
        \midrule
        \textbf{HumanEval} & \texttt{Complete the given code based on the mentioned constraints: "\textbf{humaneval\_code1}", Code: "\textbf{humaneval\_sol1}". Complete the given code based on the mentioned constraints: "\textbf{humaneval\_code2}", Code: "\textbf{humaneval\_sol1}".Complete the given "\textbf{code}" based on the mentioned constraints:  "\textbf{humaneval\_code3}", Code: "\textbf{humaneval\_sol3}".}\\
        \midrule
        \textbf{MMLU} & \texttt{Choose the answer for the question: "\textbf{mmlu\_ques1}" A. [AND, NOT] B. [NOT, OR] C. [AND, OR] D. [NAND] Answer: C. Explanation: "\textbf{mmlu\_exp1}". Choose the answer for the question "\textbf{mmlu\_ques2}" A. The defendant's statement was involuntary. B. The defendant's statement was voluntary. C. The defendant was not in custody when the statement was made. D. The statement was not made in response to a known police interrogation. Answer: A, Explanation: "\textbf{mmlu\_exp2}". Choose the answer for the question: "\textbf{mmlu\_ques3}" . A. Wrong, Wrong. B. Wrong, Not wrong C. Not wrong, Wrong D. Not wrong, Not wrong. Answer: B Explanation: "\textbf{mmlu\_exp3}". Choose the answer."\textbf{question}" ``\textbf{question}",A. ``\textbf{option 1}",B. ``\textbf{option 2}",C. ``\textbf{option 3}",D. ``\textbf{option 4}"}.\\
        \bottomrule
    \end{tabular}
    
    \label{tab:prompt_3cot}
\end{table*}
\subsection{Level 4: Least-to-Most Prompting}
Least-to-most prompting is an advanced technique that gradually increases prompt complexity, starting with simpler tasks and progressing to more complex challenges. This method allows the LLM to build confidence and leverage insights from easier prompts to tackle harder ones, enhancing its ability to generalize from straightforward examples to intricate scenarios. Table \ref{tab:prompt_lmp} displays the prompt templates used for Least-to-Most Prompting across all datasets in the experiments.
\begin{table*}[!ht]
\caption{Prompt templates of different datasets for Least-to-Most Prompting.}
    \centering    
    \begin{tabular}{p{2cm}p{12cm}}
        \toprule
        \textbf{Dataset} & \textbf{Prompt} \\
        \midrule
        \textbf{BoolQ} &  \texttt{\textbf{prompt 1}: Summarize the main points of this passage: "\textbf{passage}".
        \textbf{prompt 2}: Analyze this question to identify its key components: "\textbf{question}".\newline
        \textbf{prompt 3}: Find the part of the passage that relates to this question: "\textbf{question}", Passage: "\textbf{passage}". \newline
        \textbf{prompt 4}: Based on the passage, what is the answer to this question: "\textbf{question}", Relevant Information: "\textbf{previous response}".} \\
        \midrule
        \textbf{CSQA} & \texttt{\textbf{prompt 1}: Analyze this question: "\textbf{question}".\newline \textbf{prompt 2}: Elaborate about each option for the question: "\textbf{question}", options: A. ``\textbf{option 1}",B. ``\textbf{option 2}",C. ``\textbf{option 3}",D. ``\textbf{option 4}",E. ``\textbf{option 5}".\newline \textbf{prompt 3}: Based on the analysis: "\textbf{previous response}", discard wrong answers among the options: A. ``\textbf{option 1}",B. ``\textbf{option 2}",C. ``\textbf{option 3}",D. ``\textbf{option 4}",E. ``\textbf{option 5}".\newline \textbf{prompt 4}: Choose the correct answer from the options: A. ``\textbf{option 1}",B. ``\textbf{option 2}",C. ``\textbf{option 3}",D. ``\textbf{option 4}",E. ``\textbf{option 5}".} \\
        \midrule
        \textbf{IWSLT} & \texttt{\textbf{prompt 1}: What is the main idea or theme of this text? "\textbf{english text}".\newline \textbf{prompt 2}: Identify and list the key phrases or terms in this text: "\textbf{english text}".\newline \textbf{prompt 3}: Translate the following key phrases into French: "\textbf{previous response}".\newline \textbf{prompt 4}:
                                    Translate "\textbf{english text}" into French, incorporating the translations of the key phrases: "\textbf{previous response}".} \\
        \midrule
        \textbf{SamSum} & \texttt{\textbf{prompt 1}: List the main points or key ideas present in this dialogue: "\textbf{dialogue}".\newline \textbf{prompt 2}: Elaborate on the following key points, providing additional details or context: "\textbf{previous response}".\newline \textbf{prompt 3}: Using the listed key points and their elaborations, draft a concise summary of this text: "\textbf{dialogue}".\newline \textbf{prompt 4}: Refine this draft summary to make it more concise and coherent, ensuring it captures the essence of the text: "\textbf{dialogue}".} \\
        \midrule
        \textbf{GSM8k} & \texttt{Analyze the question: "\textbf{question}". Break the question into sub-problems: "\textbf{question}". Calculate answers for the subproblems of the question: "\textbf{pred}". Calculate the numerical answer to this question: "\textbf{question}" based on the previous calculations: "\textbf{pred}"}\\
        \midrule
        \textbf{HumanEval} & \texttt{Analyze the code: "\textbf{code}". Break the question into sub-problems: "\textbf{code}". Complete code for the subproblems of the code: "\textbf{pred}". Complete the code based on the mentioned constraints: "\textbf{code}" based on the previous calculations: "\textbf{pred}"}\\
        \midrule
        \textbf{MMLU} & \texttt{Analyze the question: "\textbf{question}". Elaborate about each option for the question: "\textbf{question}", options: A. "\textbf{option 1}" B. "\textbf{option 2}" C. "\textbf{option 3}" D. "\textbf{option 4}". Based on the analysis: "\textbf{question}", Discard wrong answers among the options: A. "\textbf{option 1}" B. "\textbf{option 2}" C. "\textbf{option 3}" D. "\textbf{option 4}"}.\\
        \bottomrule
    \end{tabular}
    
    \label{tab:prompt_lmp}
\end{table*}
\subsection{Level 5: Generated Knowledge Prompting}
Generated Knowledge prompting is one of the most complex techniques in HPF, where the LLM not only addresses the task but also integrates relevant additional information to enhance its response. This method prompts another LLM to produce auxiliary knowledge, creating a richer context for understanding and solving the problem. By leveraging self-generated insights, the LLM can deliver more detailed, accurate, and nuanced answers. Table \ref{tab:prompts_gkp} shows the prompt templates used for Generated Knowledge Prompting across all datasets in the experiments.
\begin{table*}[!ht]
\caption{Prompt templates of different datasets for Generated Knowledge Prompting.}
    \centering   
    \begin{tabular}{p{2cm}p{12cm}}
        \toprule
        \textbf{Dataset} & \textbf{Prompt} \\
        \midrule
        \textbf{BoolQ} & \texttt{\textbf{inference prompt}: Based on the passage:"\textbf{passage}", answer True/False to the question: '\textbf{question}' using knowledge of the passage:"\textbf{knowledge}"\newline \textbf{knowledge generation prompt}: Generate Knowledge about the passage: "\textbf{passage}".} \\
        \midrule
        \textbf{CSQA} & \texttt{\textbf{inference prompt}: Choose the answer:"\textbf{question}", A. ``\textbf{option 1}",B. ``\textbf{option 2}",C. ``\textbf{option 3}",D. ``\textbf{option 4}",E. ``\textbf{option 5}" using knowledge of the question:"\textbf{knowledge}" \newline \textbf{knowledge generation prompt}: Generate Knowledge about the question: "\textbf{question}".} \\
        \midrule
        \textbf{IWSLT} & \texttt{\textbf{inference prompt}: Translate "\textbf{english text}": to French using definitions of the keywords:"\textbf{knowledge}" \newline \textbf{knowledge generation prompt}: Generate definitions in french of each word in the text: "\textbf{english text}".} \\
        \midrule
        \textbf{SamSum} & \texttt{\textbf{inference prompt}: Summarize the Dialogue: "\textbf{dialogue}" using the interpretation of the dialogue:"\textbf{knowledge}" \newline \textbf{knowledge generation prompt}: Generate interpretation about the dialogue: "\textbf{dialogue}".} \\
        \midrule
        \textbf{GSM8k} & \texttt{Based on the question:"\textbf{question}", calculate the numerical answer to the question using an interpretation of the question:"\textbf{pred}"}\\
        \midrule
        \textbf{HumanEval} & \texttt{Complete the code based on the mentioned constraints:"\textbf{code}" using knowledge of the constraints:"\textbf{pred}"}\\
        \midrule
        \textbf{MMLU} & \texttt{Choose the answer. "\textbf{question}", options: A. "\textbf{option 1}" B. "\textbf{option 2}" C. "\textbf{option 3}" D. "\textbf{option 4}" using knowledge of the question:"\textbf{pred}"}\\
        \bottomrule
    \end{tabular}
    
    \label{tab:prompts_gkp}
\end{table*}

\end{document}